\documentclass[twocolumn, switch]{article} 
\usepackage{styles/preprint}

\usepackage{amsmath, amsthm, amssymb, amsfonts}

\usepackage[numbers,square]{natbib}
\usepackage[utf8]{inputenc}	
\usepackage[T1]{fontenc}	
\usepackage{xcolor}		
\usepackage{hyperref}	
\usepackage{booktabs} 		
\usepackage{nicefrac}		
\usepackage{microtype}		
\usepackage{lineno}		
\usepackage{float}			

\usepackage{diagbox}
\usepackage{graphicx}
\graphicspath{{figures/}}
\DeclareGraphicsExtensions{.pdf,.jpeg,.png}
\usepackage[caption=false,font=normalsize,labelfont=sf,textfont=sf]{subfig}

\usepackage{newfloat}
\DeclareFloatingEnvironment[name={Supplementary Figure}]{suppfigure}
\usepackage{sidecap}
\sidecaptionvpos{figure}{c}

\usepackage{titlesec}
\titlespacing\section{0pt}{12pt plus 3pt minus 3pt}{1pt plus 1pt minus 1pt}
\titlespacing\subsection{0pt}{10pt plus 3pt minus 3pt}{1pt plus 1pt minus 1pt}
\titlespacing\subsubsection{0pt}{8pt plus 3pt minus 3pt}{1pt plus 1pt minus 1pt}

\newcommand{\grid}{\mathcal{G}}
\newcommand{\gridpoint}{g_k}
\newcommand{\cell}{V_k}
\newcommand{\tensor}{T}
\newcommand{\model}{{\bf m}_i}
\newcommand{\val}{{v}_j}
\newcommand{\pos}{{\bf p}_j}
\newcommand{\dir}{{\bf d}_j}

\newcommand{\subS}{S_i}
\newcommand{\function}{F_{i}}
\newcommand{\cellsubS}{S_{ik}}
\def\eg{\textit{e.g.}\;}
\def\etal{\textit{ et al.}\,}
\def\ie{\textit{i.e.}~}
\def\cf{\textit{cf.}\;}

\newcommand{\real}{\mathbb{R}}

\newcommand{\clearemptydoublepage}{%
  \ifthenelse{\boolean{@twoside}}{\newpage{\pagestyle{empty}\cleardoublepage}}%
  {\clearpage}}
	
\newcommand{\meanStd}[2]
 { #1\,$\pm$\,#2}

\newcommand{\shortcaption}[1]
  {\textbf{#1}}
	
\newcommand{\mysubref}[1]
  {\textbf{(#1)}}

\title{Redefining Ultrasound Compounding: Computational Sonography}

\usepackage[blocks]{authblk}

\author[*,1]{R\"udiger G\"obl}
\author[*,1]{Diana Mateus}
\author[1]{Christoph Hennersperger}
\author[1]{Maximilian Baust}
\author[1]{Nassir Navab}
\affil[1]{Computer Aided Medical Procedures, Technische Universit\"at M\"unchen, Boltzmannstr. 3, 85748 Garching, Germany.}
\affil[*]{R.~G\"obl and D.~Mateus contributed equally to this work.}

\begin{document}

\twocolumn[ 
  \begin{@twocolumnfalse} 
  
\maketitle

\begin{abstract}
Freehand three-dimensional ultrasound (3D-US) has gained considerable interest in research, but even today suffers from its high inter-operator variability in clinical practice. 
The high variability mainly arises from tracking inaccuracies as well as the directionality of the ultrasound data, being neglected in most of today's reconstruction methods. 
By providing a novel paradigm for the acquisition and reconstruction of tracked freehand 3D ultrasound, this work presents the concept of Computational Sonography (CS) to model the directionality of ultrasound information. 
CS preserves the directionality of the acquired data, and allows for its exploitation by computational algorithms.
In this regard, we propose a set of mathematical models to represent 3D-US data, inspired by the physics of ultrasound imaging.
We compare different models of Computational Sonography to classical scalar compounding for freehand acquisitions, providing both an improved preservation of US directionality as well as improved image quality in 3D.
The novel concept is evaluated for a set of phantom datasets, as well as for in-vivo acquisitions of muscoloskeletal and vascular applications.

\end{abstract}
\vspace{0.35cm}

  \end{@twocolumnfalse} 
] 



\section{Introduction}
\label{sec:intro}

	In recent years, 3D ultrasound (US) imaging has  gained interest~\cite{brattain:ipcai2012,neshat:medical-physics2013}  both due to the greater availability of 3D ultrasound systems (tracked, mechanical driven or matrix arrays)\cite{fenster:intech2013,vegas:aa2010} as well as to the improvements in computational power allowing for faster and online processing of the acquisitions. To date, the most common way to obtain a 3D US volume remains the interpolation of 2D images into a 3D volume~\cite{solberg:usmb2007}.
We focus here on {\it tracked 3D freehand ultrasound}, a technique that yields 3D US volumes by first acquiring 2D images along with their position and orientation and then fusing them into a volumetric regular grid through a process known as 3D reconstruction or \textit{compounding}~\cite{solberg:usmb2007}. 
Up to now, conventional compounding techniques (\eg forward, backward or functional interpolation) have neglected an essential property of ultrasound images, namely, its direction-dependent nature. 
It is known that imaging a point from different directions may lead to different intensities. 
This direction-dependency is caused by the different attenuation paths followed by the reflected signals, as well as by the dependency of the reflected intensity on the tissue's impedance and orientations as defined by Snell's law.
This directional behavior in part explains the information loss resultant from current 3D reconstruction methods\cite{prager:ultrasonics2002} and is the main reason for restricted acquisition protocols (\eg straight probe motion) recommended in practice. It is to be noted that such protocols modify the physician's common practice based on image feedback and interactive motion of the probe.

\begin{figure}
	\begin{center}
		\includegraphics[width=\linewidth,trim=4cm 3cm 4cm 4cm,clip]{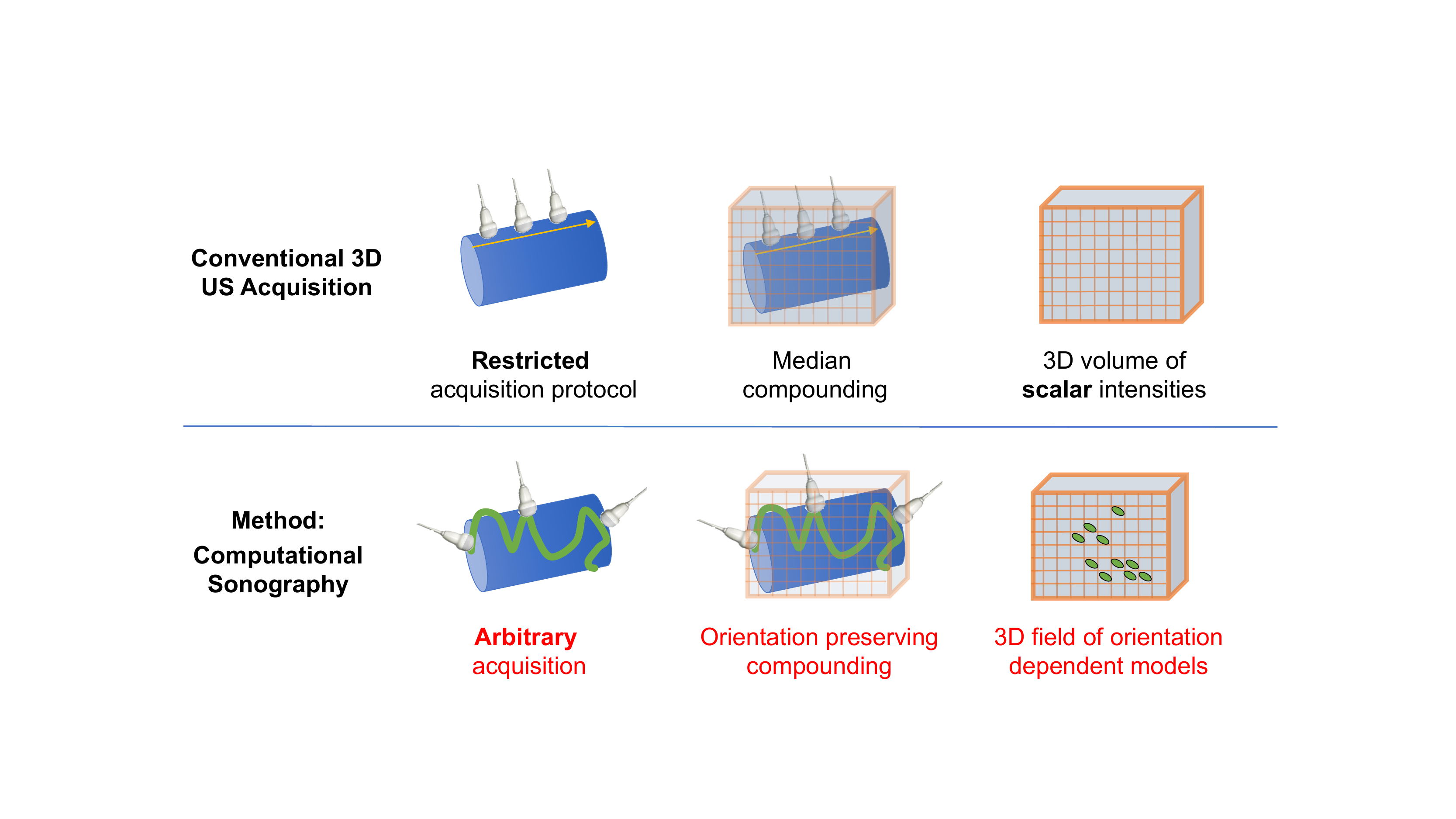}
	\end{center}
	\caption{\shortcaption{Computational Sonography concept.}}
	\label{fig:overview}
\end{figure}

In this work, we introduce \textit{Computational Sonography} (CS) as a novel paradigm for the {\it 3D reconstruction and representation} of tracked freehand 3D US that opens new possibilities for visualizing, interacting and processing volumetric data. In more detail, we introduce  two major changes to current reconstruction paradigms (\cf \autoref{fig:overview}). First, we consider and preserve the directional dependency of US in the 3D reconstruction. And second, we enable each element in the 3D volume to store a complex model instead of just a single scalar value.
We argue that with such complex models it becomes possible to retain the directional information previously discarded and make it vailable for further use. 
We want to point out that the ``Computational'' term in CS, is an analogy to the concept of Computational Photography~\cite{Ng05}, and stands for the fact that once such a more complex volumetric representation has been stored, we deploy computational algorithms in order to extract and display the relevant information on demand. 

\begin{figure}
\begin{center}
\subfloat[]{\includegraphics[width=0.31\linewidth,trim=0  1.1cm 0 0.5cm,clip]{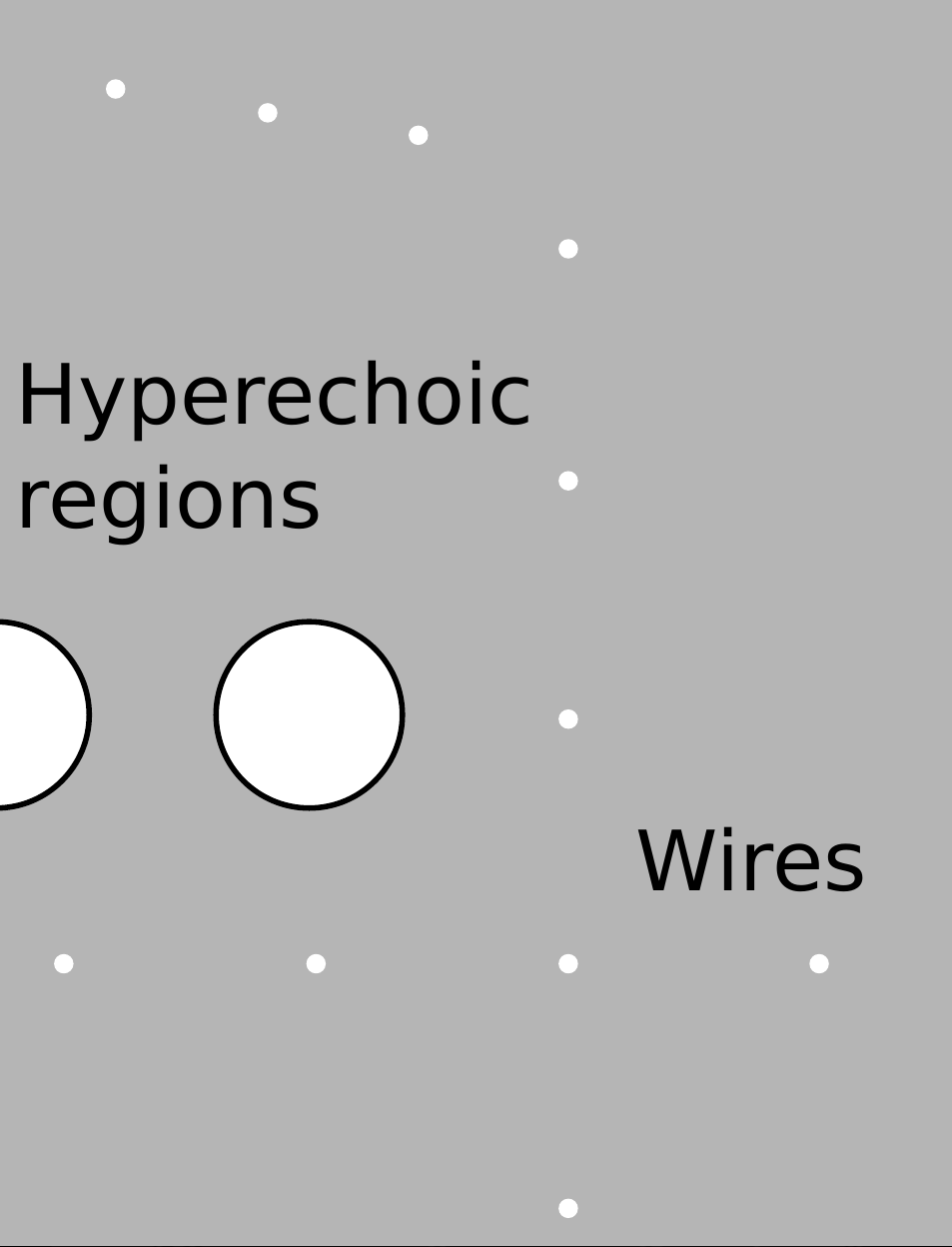}\label{fig:directionalityProblem3}}
\hspace{3pt}
\subfloat[]{\includegraphics[width=0.31\linewidth,trim=0  1.1cm 0 0.9cm,clip]{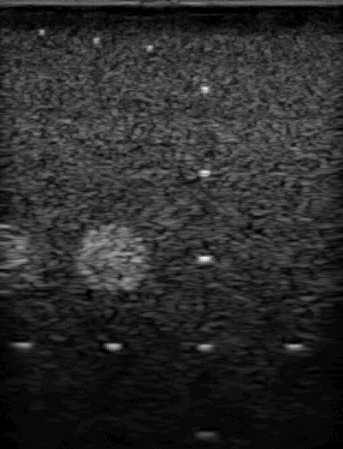}\label{fig:directionalityProblem1}}
\hspace{3pt}
\subfloat[]{\includegraphics[width=0.31\linewidth,trim=0 1.1cm 0 0.9cm,clip]{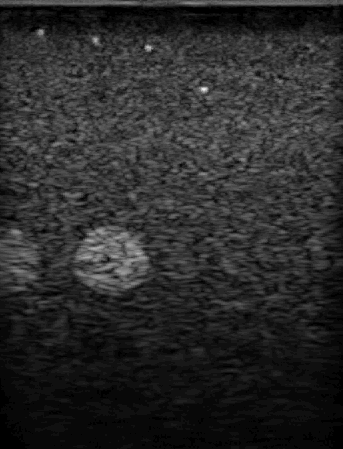}\label{fig:directionalityProblem2}}

\end{center}
\caption{Illustration of the \shortcaption{directionality dependency in US imaging} using images of a phantom depicted in 
    \mysubref{a}.
    \mysubref{b} When the probe is perpendicular to the wires these appear clearly in the image.
    \mysubref{c} When the probe is tilted the wires may not be visible.
    }
\label{fig:intro}
\end{figure}

The goal of this paper is to demonstrate how the concept of Computational Sonography can be translated into new direction-preserving compounding methods. 
In practice, we present two methods that reconstruct the amount of reflected signal expected from every direction at each point in a volume. 
An important design specification is the model to represent these values, where desired properties are the directionality-preservation but also compactness, and fixed-length per voxel. 
We study two representation models complying with the requirements:  tensors, as initially explored in a conference paper~\cite{hennersperger2015computational}, and a novel spherical representation.
In summary, our contributions to 3D sonography are:
\begin{itemize}
\item A reconsideration of tacit assumptions in the process of fusing multiple US views regarding i) the directionality of US, and ii) the reconstruction of scalar volumes.
\item The introduction of two models to represent multi-directional data, namely tensor-CS and spherical-CS.
\item The analysis of two direction-preserving compounding algorithms and their comparison to classical compounding.
\item The demonstration of key capabilities of CS.
\end{itemize}

Finally, to show the practical value, we quantitatively evaluate the fidelity of the CS reconstructions (vs. information loss).
The qualitative and quantitative evaluations are performed for both phantom and in-vivo data in comparison to state-of-the-art compounding methods.

This work follows an initial conference publication~\cite{hennersperger2015computational} but includes the following extensions: in addition to the tensor model we propose a more expressive spherical representation, which encodes the directionality of ultrasound-information with minimal information-loss. 
We also further analyse in-depth and quantitatively compare the two models against conventional compounding methods for both phantom and in-vivo acquisitions. 
In particular, we have extended the evaluation of the fidelity from 15 in-vivo and in-vitro cases to 40 cases.

\subsection{Related Work}
\label{sec:related}

Only few  methods have considered the directionality information in the processing 3D US volumes. Efforts in our group include a method to solve quadratic optimization problems in freehand 3D US in a compounding-free fashion~\cite{hennersperger:miccai2014}, as well as the design of directional dependent filters to vessels in 3D US images~\cite{hennersperger:tmi2015}. For 3D multiple-view US reconstruction, Schulte zu Berge\etal\cite{zu2014orientation} use the orientation information to cluster 2D slices, compound one volume per cluster and fuse the clusters. 
Different to CS, this method reconstructs only a scalar field, the directional information is lost for further processing.

Towards reconstructing more complex models per voxel, Klein\etal\cite{klein2012modeling} cluster the samples in a voxel according to their viewing angles but combine them into a volume of mixture models of Nakagami distributions, resulting in a vector containing the distribution parameters per voxel. However, the directionality information is also lost after reconstruction.
Recently, Lee\etal\cite{lee2012ultrasound} and Papadacii\etal\cite{papadacci2014ultrasound} proposed tensorial-based reconstructions (respectively Elastic Tensor Imaging (ETI) and Backscatter Tensor Imaging (BTI)) of US images from multiple view angles. 
Based on shear wave imaging, ETI measures the anisotropy of tissue elasticity and represents it with a tensor.
In BTI, the local spatial coherence of the US signals observed for different orientations and focal regions was used to reconstruct 3D tensor volumes.
Although these methods also recover tensor fields, their focus is on recovering the fiber orientation of the underlying tissues. Additionally, they both rely on a specialized hardware setup and a fixed scanning protocol. In contrast, our approach focuses on information preservation of the full acquisition and can be seamlessly integrated into the clinical workflow as well as applied directly to clinical 3D freehand US acquisitions.
Finally, beyond tensor fields, we have also proposed a new representation for 3D US, Spherical-CS.

Representing US directional dependent information is similar to the problem of describing diffusion patterns in Diffusion MRI. 
In this field, both tensors (Diffusion Tensor Imaging-DTI\cite{le2001diffusion}) and spherical representations (Q-ball\cite{tuch:MRIm2004:qball}) have been studied. 
However, apart from the representation model, the methods to fit and visualize tensors and q-balls are specialized to the physics of MRI acquisition and therefore cannot be applied in a straightforward manner to 3D US data. 

In conclusion and to the best of our knowledge, this is the first orientation preserving reconstruction method for 3D ultrasound. The proposed CS-models bear some similarity to those used in diffusion MRI, but their computation from ultrasound data is new.

\section{Computational Sonography Concept}
\label{sec:problem}

We investigate the problem of building a {\it direction-dependent 3D US volumetric reconstruction} from a set of tracked 2D US images consisting of a set of tracked US samples\footnote{US images are composed of rays, each formed by {\it samples} resulting from  measuring the reflected signal at regular times.}. We denote the set of acquired samples  $S=\{s_j\}_{j=1}^{N_{\rm samp}}$ and perform the reconstruction on a 3D volumetric grid in a lattice $X=\{{\bf x}_i\}_{i=1}^{N_{\rm vox}}$, with ${\bf x}_i\in \mathbb{Z}^3$.
We can then decompose the reconstruction problem in two tasks: 
\begin{enumerate}
\item Finding the relevant subset of samples $\subS\subset S$ contributing to the reconstruction of each voxel ${\bf x}_i$.
\item Fitting the chosen samples $\subS$ to the orientation-dependent model ${\bf m}_i$ stored in each voxel ${\bf x}_i$.
\end{enumerate}

While we follow the strategy of \cite{hennersperger2014ipcai} for the first stage (\cf \autoref{sec:sample-selection}), we  focus hereafter on the reconstruction task.
The problem can be formalized as finding a reconstruction function $R$ mapping from the set of measured samples $\subS$ to a  model  $\model$ for every voxel in $X$, 
\begin{equation}
R: S_i \mapsto \model.
\end{equation}

Let each US sample $s_j$ be characterized by a  tuple $ (\val, \pos, \dir)$, where $\val$ is the measured reflected signal, and $\pos \in \real^3$ and $\dir \in \mathbb{S}^2$ stand respectively for the sample's position and acquisition direction. 
In conventional compounding methods, $R$ maps the samples to a scalar value $\model=m_i\in\real$.
Accordingly, the mean reconstruction can be formalized as 
\begin{eqnarray}
\label{equ:compoundingMean}
R_{\rm mean}: \subS \mapsto  &\frac{1}{|S_i|}\sum_{j=1}^{|S_i|} v_j.
\end{eqnarray}
Other common scalar reconstructions include the median, k-Nearest Neighbor or Inverse Distance Weighted compounding methods.

With computational sonography, we propose two new reconstruction functions $R$, which fuse samples $S_i$ while preserving ultrasound's directionality; \ie they capture the fact that the reconstruction may show different intensity values for different directions.
The two proposed methods, one tensor-based $R_{T}$ and another spherical-based $R_{\rm Sph}$, differ in the fusion as well in the final model used for compact, but direction preserving, representation $\model$ of the reconstruction, as detailed in  \autoref{sec:methodology}.

The result of the two CS reconstruction procedures is a compact representation of the full acquisition, which preserves the directional nature of US for visualization, as well as other possible common post-processing tasks.

\section{Methodology}
\label{sec:methodology}
In order to design a direction-preserving reconstruction function $R$, we propose to store a representation model $\model$ instead of a scalar value in each element of a target lattice.
We present two reconstruction functions and representation models. The first, initially introduced in \cite{hennersperger2015computational}, is a tensor-based approach, Tensor Computational Sonography (Tensor-CS). The second, is a novel approach relying on the concept of spherical grids (Spherical-CS).

\subsection{Tensor-Models}
\label{sec:tensors}
For Tensor-CS, we choose to fill the output volumetric grid with tensors, that is we store a tensor model $\model = T_i\in\real^{3\times 3}$ in each voxel. 
The reconstruction function $R$ of Tensor-CS is implicitly defined as:
\begin{equation}
R_{T}: \subS \mapsto \tensor_i \in \mathbb{R}^{3 \times 3},\; {\rm s.t.}\; \val = \dir ^\top \tensor_i \dir,
\end{equation}
where the tensor $\tensor_i$ is assumed to be symmetric.
The direction preservation is achieved by means of the constraint $\val = \dir ^\top \tensor_i \dir$, which states that tensor $\tensor_i$ should compactly encode the intensity value $\val$ observed from the given direction $\dir$. 
In practice, a least squares approach is employed to find the values of the tensor $\tensor_i$, where we minimize
\begin{equation}
\textrm{min}_{\tensor_i} \sum_{j=1}^{|\subS|} \left\|\dir ^\top \tensor_i \dir - \val\right\|_2^2.
\end{equation}
As $T_i$ shall be symmetric, at least six sampling points determine the tensor coefficients.
In practice, we use  typically between 12 and 500 samples, as this showed to provide best results in our experiments.
We solve the least squares problem using Cholesky decompositions implemented on a GPU.

\begin{figure*}
	\captionsetup[subfigure]{labelformat=empty}
	\setlength{\tabcolsep}{0.8em}
	\begin{tabular}{cccc}
		\subfloat[][(a)]{\includegraphics[width=0.26\textwidth]{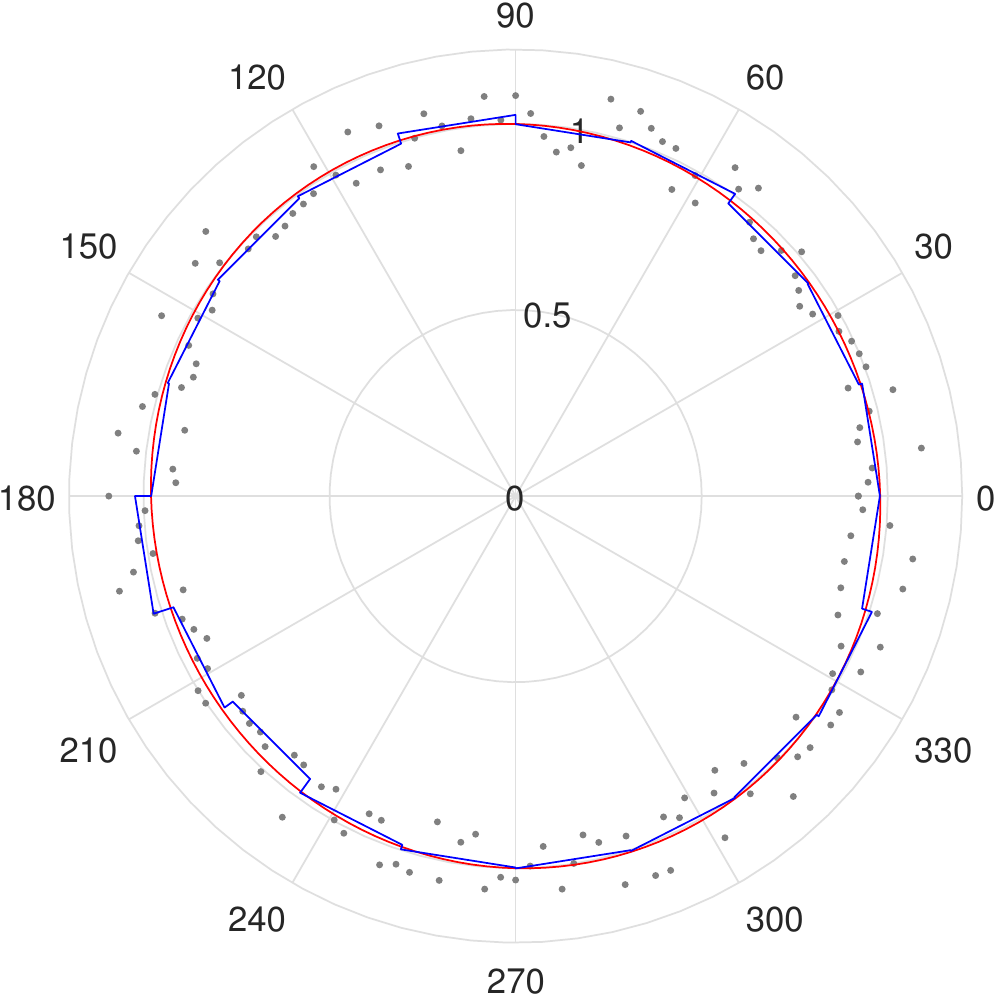}\label{fig:shadowing-a}} &
		\subfloat[][(b)]{\includegraphics[width=0.26\textwidth]{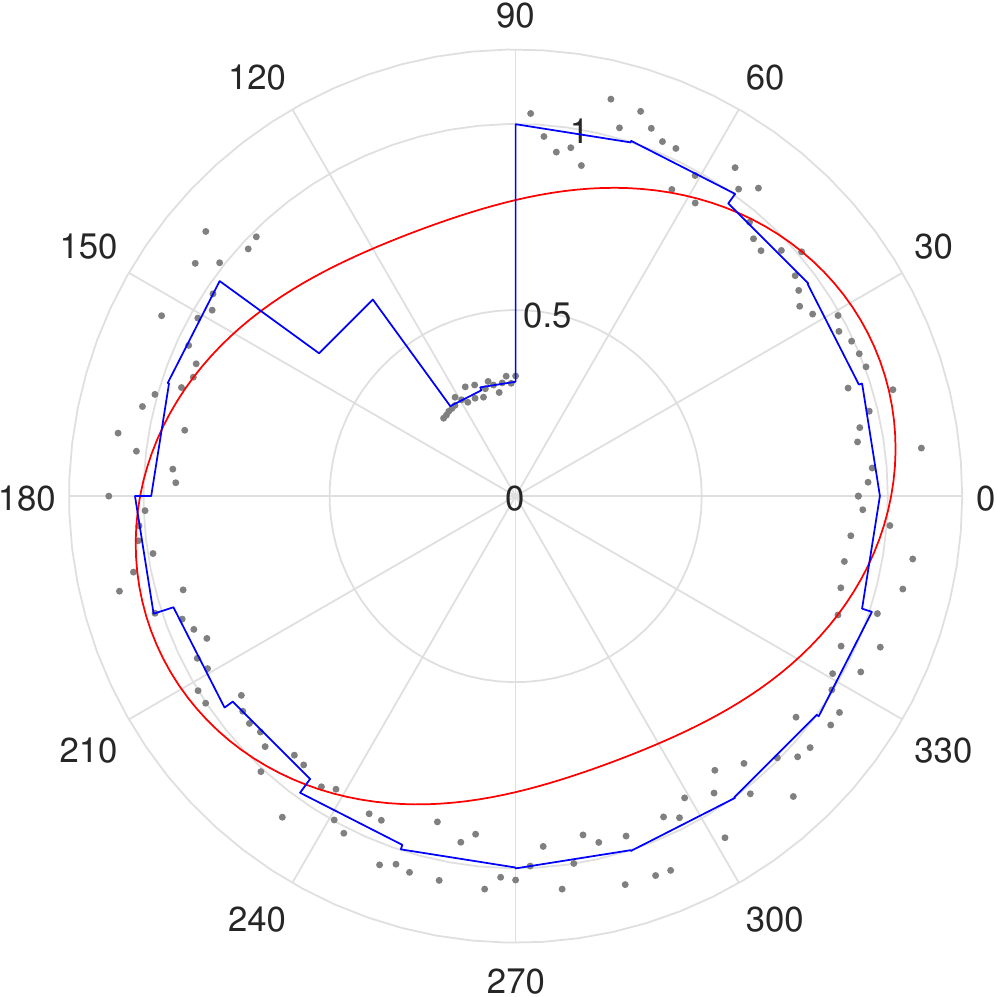}\label{fig:shadowing-b}} &
		\subfloat[][(c)]{\includegraphics[width=0.26\textwidth]{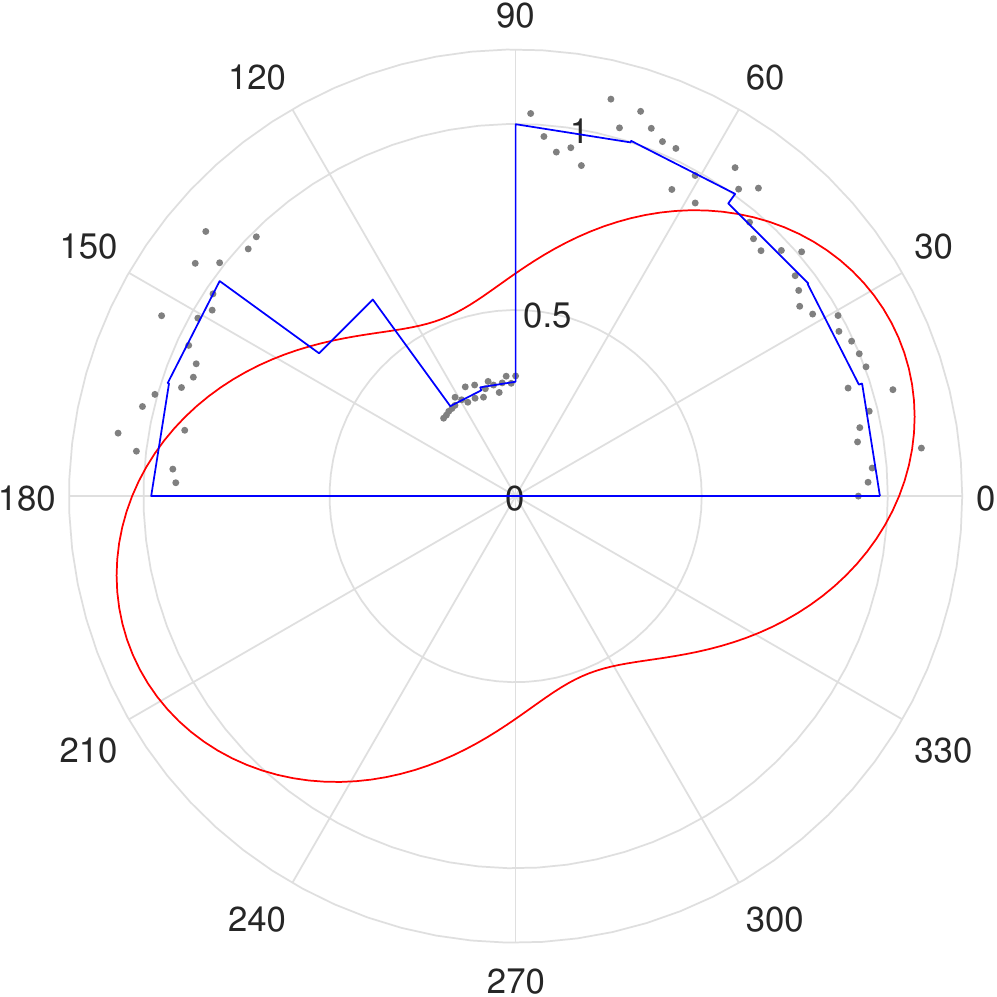}\label{fig:shadowing-c}} &
		\subfloat{\includegraphics[width=0.15\textwidth,angle=90]{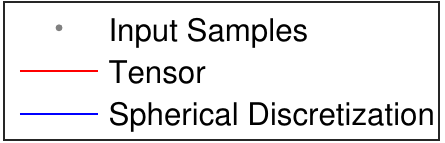}}
	\end{tabular}
	\caption{\shortcaption{Symmetry and two dimensional Computational Sonography}. We simulate two distributions of measured samples, shown as gray dots. \mysubref{a} The first distribution is point-symmetric. \mysubref{b} In the second distribution the symmetry is broken by a ``shadow". In \mysubref{c}, only  measurements from a restricted angle are considered.
The reconstruction with {\bf Tensor-CS (red)} fits nicely to noisy point-symmetric data \mysubref{a}, but has problems with dissymmetries, as they could result from shadowing \mysubref{b}.
These artefacts are somewhat mitigated when acquisitions are taken only from a restricted angle as in \mysubref{c}.
The reconstruction obtained with {\bf Spherical-CS (blue)} accurately approximates the signal in the three cases, as shown by the blue lines from \mysubref{a} to \mysubref{c}.
\label{fig:shadowing} }
\end{figure*}

The main advantage of the tensor model is that it allows for a direction dependent modeling of the ultrasound signal while requiring only six real values to be stored per voxel, making Tensor-CS very suitable for parallel processing. The tensor model is also suited for an interactive orientation-dependent visualization (see \autoref{sec:computational}). Finally, tensor fields can be further post-processed by means of tensor-based regularization and visualization methods, such as those used in DTI \cite{arsigny2005fast, christiansen2007total, rosman2014fast}.
On the other hand, the tensors' compactness comes at the price of  resolution loss and some simplifications. 
For instance,  the tensor modeling assumes that signals are point symmetric. While the symmetry is irrelevant for the most common freehand acquisitions, it becomes a problem for acquisitions from all orientations, and in particular, in the presence of acoustic shadowing. This phenomenon is illustrated in \autoref{fig:shadowing-a} and \autoref{fig:shadowing-b}. 

\subsection{Spherical Models}
\label{sec:spheres}
To overcome some of the limitations of Tensor-CS regarding the resolution and enforced symmetry, we propose a new representation for 3D US data.
More precisely, for each voxel we model the ultrasound signal reflected from all directions as a continuous real function $F$ defined on the unitary sphere $F:\mathbb{S}^2\rightarrow \real$. This model allows a unique intensity value $v$ for every direction ${\bf d}$, or equivalently,  $F: {\bf d}\mapsto v$.  As a result, the Spherical-CS representation consists of one spherical function per element of the lattice $X$, that is, $\model=\function$.
Recall that the samples $\{s_j\}\in S_i$ relevant to voxel $i$  are defined by $s_j=(\val, \pos, \dir)$, \ie by their intensity $\val$, position $\pos$ and imaging direction $\dir$. We then try to approximate the spherical function $\function$ from the samples' tuples $\{\val, \dir\}$.
Although one could store every sample relevant to each voxel to approximate $\function$  with the highest fidelity, the resulting representation would be very inefficient, due to a varying number of samples.
Instead, we choose as domain a spherical grid, computing a discrete approximation $\tilde{\function}$ of $\function$. The domain is defined as a partitioning the sphere into $N_{\rm cells}$ segments $\cell$, such that $\mathbb{S}^2=\bigcup\{\cell\} \textrm{ with } k=1,\ldots,N_{\rm cells}$. Then, we approximate a scalar value $\tilde{\function}(k)$ per cell. Finally, we collect the values from the different cells to store a model $\model = \tilde{\function}\in\real ^{N_{\rm cells}} $ in each voxel. 
Thereby, the sought reconstruction function is:
\begin{eqnarray}
R_{\rm Sph}:  \subS \mapsto \tilde{\function}\in \real ^{N_{\rm cells}}.
\end{eqnarray}
In practice, each entry of the vector $\tilde{\function}(k)$ is computed only from the samples falling inside the corresponding cell $\cell$. First we determine the subset of samples $\cellsubS\subseteq \subS$ associated to each segment $\cell$. Then, the value $\tilde{\function}(k)$ assigned to the cell $\cell$ is obtained from the samples intensity values $\val$ through one interpolation schemes of
\begin{eqnarray}
R_{\rm Sph-mean}: \cellsubS\mapsto&\tilde{\function}(k) = \frac{1}{|\cellsubS|}\sum_{j=1}^{|\cellsubS|} \val,\label{eq:recofirst}\\ 
R_{\rm Sph-med}: \cellsubS \mapsto & \tilde{\function}(k) = \text{median}(\val),
\end{eqnarray}
where $|\cellsubS|$ stands for the number of samples in segment $k$.
They are derived from classical compounding methods, but applied only to the associated sample subset $\cellsubS$

This discretization scheme permits adjusting the desired resolution of the representation. 
In this way, we achieve a compact, yet accurate representation of the acquisition allowing for subsequent processing or extraction of information according to the needs of the user. 
The proposed models are also well capable of representing arbitrary functions including non-point-symmetric data (\cf \autoref{fig:shadowing}).

A challenge with the spherical representation arises from the discretization of the sphere, which is itself an open subject of research~\cite{staniforth2012horizontal}.
We are interested in a discretization that:
i) leads to a regular partition of the area on the sphere, 
and ii) enables computational efficient mappings from a directions ${\bf d} \in \mathbb{S}^2$ to the respective segment $\cell$ and backwards, as these mappings will be needed for the reconstruction as well as for later access, processing or visualization.

Next, we analyse and discuss three possible discretization approaches, as well as their mappings. 
To this end, we first define a grid on the unit sphere as a set of points $\grid=\{\gridpoint\}_{k=1}^{N_{\rm cells}}$, where each point $\gridpoint: \real^2\rightarrow\mathbb{S}^2$ is described by its spherical coordinates, namely the azimuthal and elevation angles, \ie $\gridpoint=(\theta_k,\phi_k)$. 
Then, the Voronoi cells $\cell$ of the grid points inherently define a partition of the sphere.

\subsubsection{Latitude-longitude grids}
Given a sphere parametrized with spherical coordinates, the latitude-longitude discretization is simply obtained through an equidistant sampling of the azimuth and elevation angles.
As a result, it is trivial to map direction vectors to their respective point on the spherical grid by converting Euclidean coordinates to spherical coordinates and visce-versa.
More precisely, the coordinates of the points on a latitude-longitude grid $\grid^\text{LatLong}$ with a resolution of $r$ radians are given in spherical coordinates by
\begin{align}
\grid^\text{LatLong} = &\{(\theta_k,\phi_k)\}_{k=1}^{N_{\rm cells}}, \\
\theta_k \in & \left\{0, \tfrac{\pi}{r}, 2\tfrac{\pi}{r}, \ldots, \pi\right\}, \\
\phi_k \in &\left\{0, \tfrac{2\pi}{r}, 2\tfrac{2\pi}{r}, \ldots, (r-1)\tfrac{2\pi}{r}\right\}.
\end{align}

As it is apparent in \autoref{fig:spherical_grid_latlong}, the inconvenience of latitude-longitude grids is that the density of points varies  significantly across the surface of the sphere, which results in cells of variable area, and is undesired for computational sonography.

\subsubsection{Icosahedral geodesic grids}
Geodesic grids are built by iteratively subdividing a polyhedron.
It is common to use a icosahedron as a starting point.
At every iteration, each polygon of the current polyhedron is further divided in cells, and projected onto the sphere.
This subdivision is continued until the desired resolution is reached.
Following this discretization, the obtained partitions have near-equal area.
For a detailed description we refer to the work of Thuburn \cite{Thuburn1997}.
\autoref{fig:sperical_grid_icosahedron} shows a geodesic grid based on an icosahedron subdivision.
Similar to the latitude-longitude grids, the spherical coordinates of the grid-points are known by construction and the mapping to their represented directions is trivial. 
However, no constant-time mapping of arbitrary direction vectors to the corresponding grid-point (or directional segment) exists for these grids \cite{keinert2015spherical}.
Since measurements can originate from arbitrary directions, icosahedral geodesic grids are not practical from a computational point of view.

\subsubsection{Fibonacci geodesic grids}
\begin{figure*}
\begin{center}
    \subfloat[]{\includegraphics[height=0.24\textwidth]{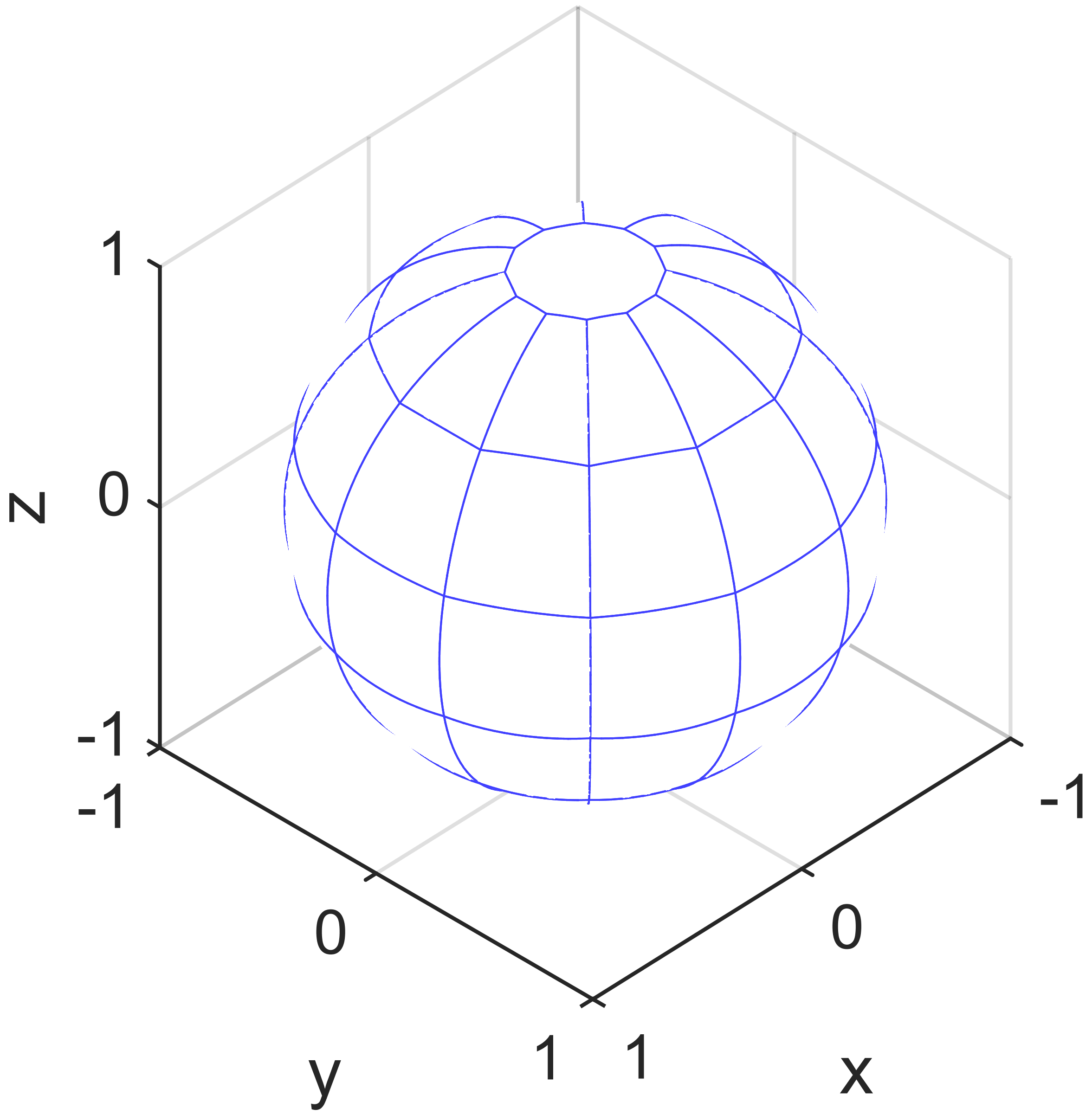}\label{fig:spherical_grid_latlong}}
    \subfloat[]{\includegraphics[height=0.24\textwidth]{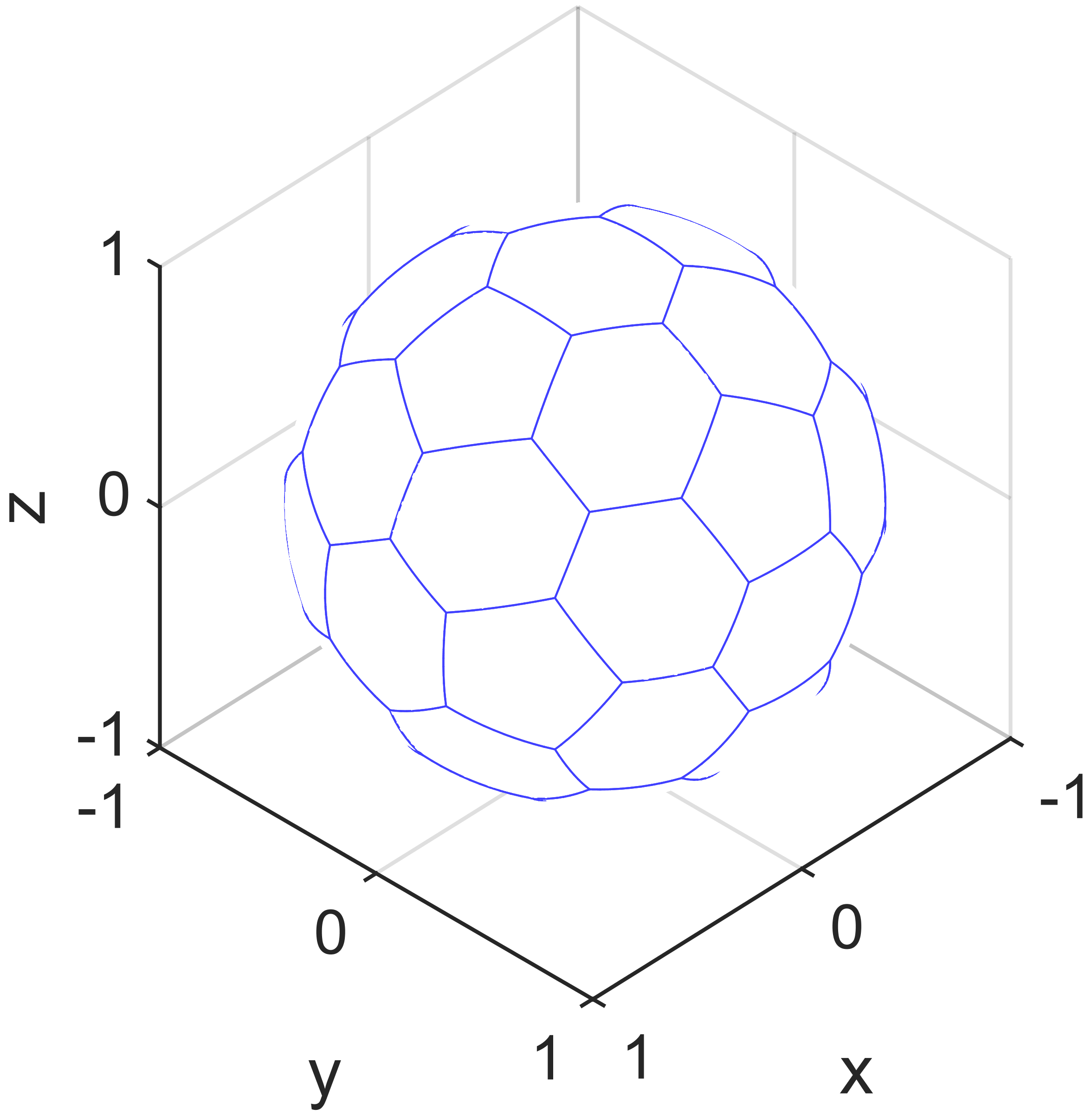}\label{fig:sperical_grid_icosahedron}}
    \subfloat[]{\includegraphics[height=0.24\textwidth]{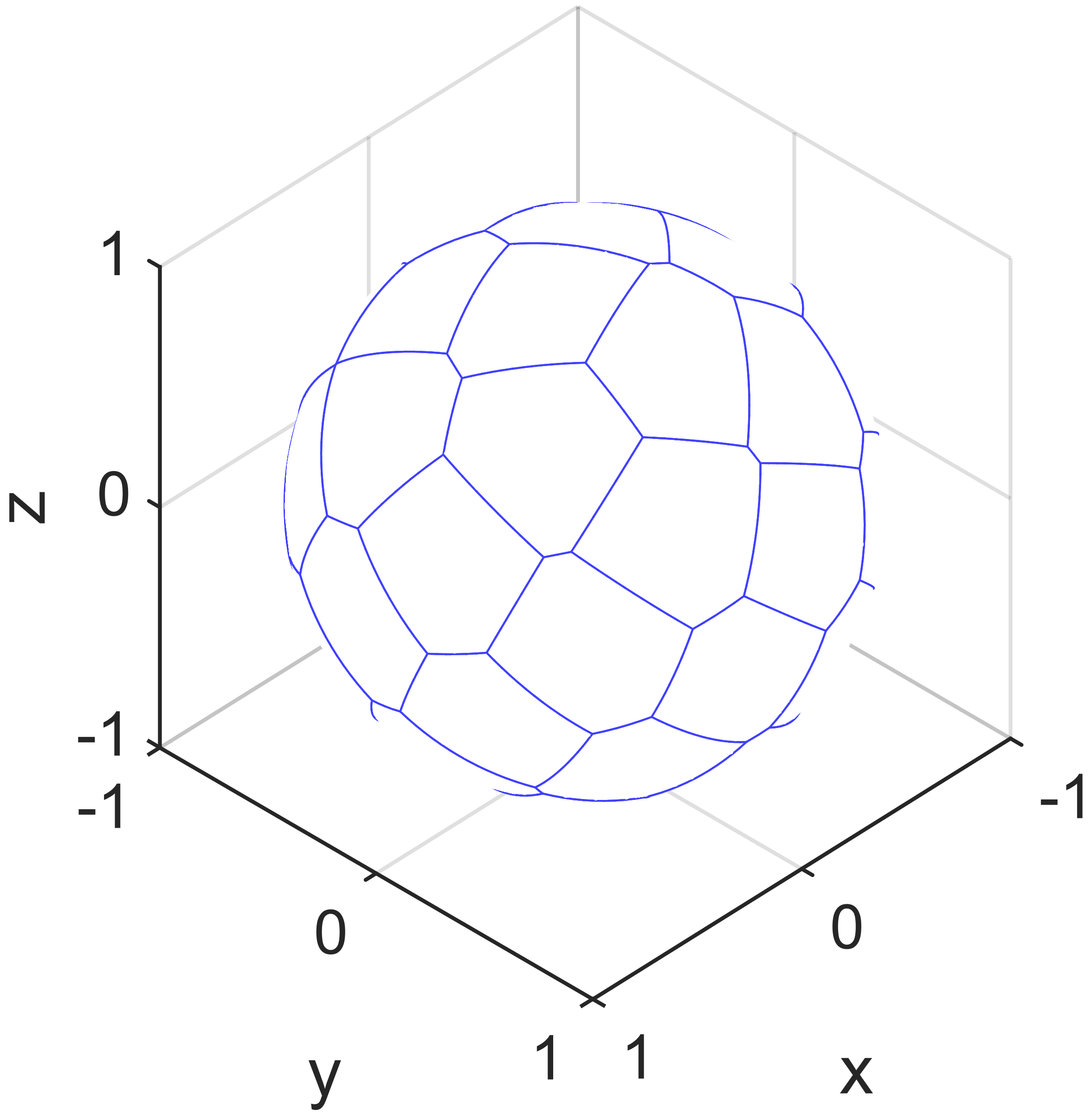}\label{fig:sperical_grid_fibonacci}}
    \subfloat[]{\includegraphics[height=0.24\textwidth]{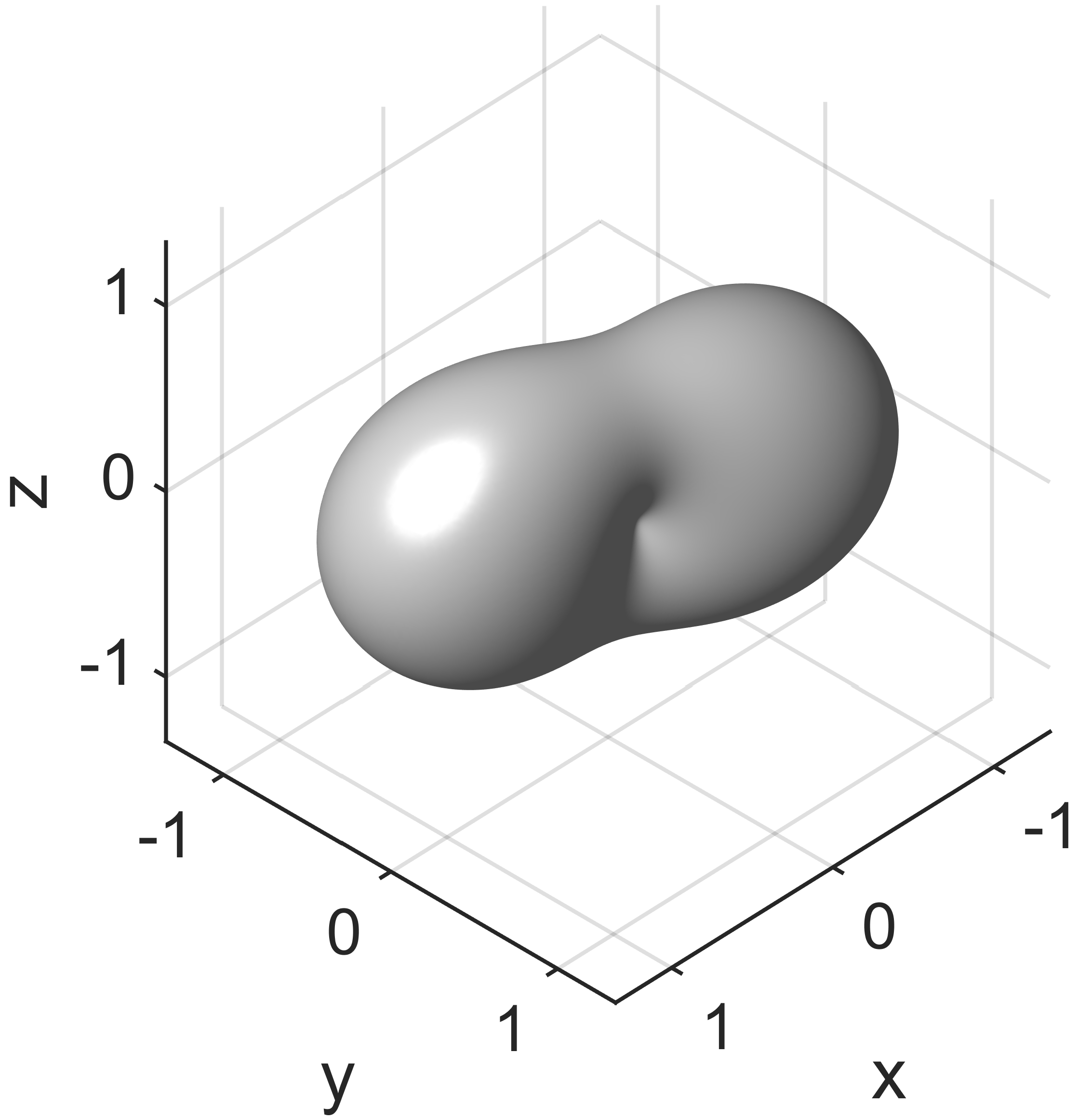}\label{fig:example_tensor}}
    \caption{\shortcaption{Voronoi Diagrams} \mysubref{a}-\mysubref{c} showing the extent of each cell in the partition of the sphere.
    \mysubref{a} Latitude-longitude grid with 30 degree resolution.
    \mysubref{b} Icosahedral spherical grid with 42 grid-points. 
    \mysubref{c} Spherical Fibonacci grid with 42 grid-points.
    \mysubref{d} \shortcaption{Exemplary tensor} evaluated in all directions.
    }
    \end{center}
\end{figure*}

The third type of discretization we studied here are spherical Fibonacci grids, which are generated by drawing points along a spiral on the surface of the sphere. The construction rules describe each grid-point in terms of the golden ratio $\Phi$ \cite{Swinbank2006}, as follows:
\begin{multline}
	\grid^\text{Fibonacci} = \{(\phi_k, cos^{-1}(z_k))\}_{k=0}^{N_{\rm cells}-1}, \\
	~ \phi_k = 2 \pi \frac{k}{\Phi} ,\; z_k = 1 - \frac{2k + 1}{N_{\rm cells}}, k \in \{0,\ldots,N_{\rm cells}-1\}.
\end{multline}
\autoref{fig:sperical_grid_fibonacci} shows a spherical Fibonacci grid with 42 grid-points.
Mapping a grid-point to a direction vector is equivalent to converting spherical to Euclidean coordinates.
Most importantly, however, the inverse mapping recently proposed by Keinert\etal\cite{keinert2015spherical} provides a mapping from any direction vector to the nearest grid-point or directional segment with constant run-time.
We refer the reader to the work for the derivation of the inverse mapping.
This property makes Fibonacci geodesic grids the most suited for computational sonography.

\subsection{Sample Selection}
\label{sec:sample-selection}
As discussed in \autoref{sec:problem}, the compounding process is composed of two stages. So far we focused on the reconstruction step and assumed the subset of measurements relevant to fit one voxel model is known. This subset is determined using a sample selection strategy initially described in~\cite{hennersperger2015computational}, and inspired by \cite{hennersperger2014ipcai}. For completeness, we describe the process next.

The task is to select all samples $s_j \in S_i\subseteq S$ which contribute to the reconstruction of a specific voxel, centered at coordinates ${\bf x}_i=[x_{i,x}\;\;x_{i,y}\;\; x_{i,z}]^\top$. Practically, we select a sample $s_j$ if the voxel center ${\bf x}_i$ lies inside the region of influence of the ultrasound sample. This region is defined as an ellipsoid centered around the sample's position $\pos=[p_{j,x}\;\;p_{j,y}\;\;p_{j,z}]^\top$ and with one of its axis aligned to the sample's direction $\dir$. Formally, $s_j\in S_i$ if:
\begin{equation}
\tfrac{1}{a^2}(x_{i,x}-p_{j,x})^2 +\tfrac{1}{b^2}(x_{i,y}-p_{j,y})^2 + \tfrac{1}{c^2}(x_{i,z}-p_{j,z})^2 \leq 1,
\label{equ:ellipsoidSelection}
\end{equation}
where $a,b,c$ represent the half of the length of the ellipsoid's axes in each direction.
\begin{figure}
\begin{center}
	\includegraphics[width=0.6\linewidth,clip,trim=0 2cm 0 1cm 0]{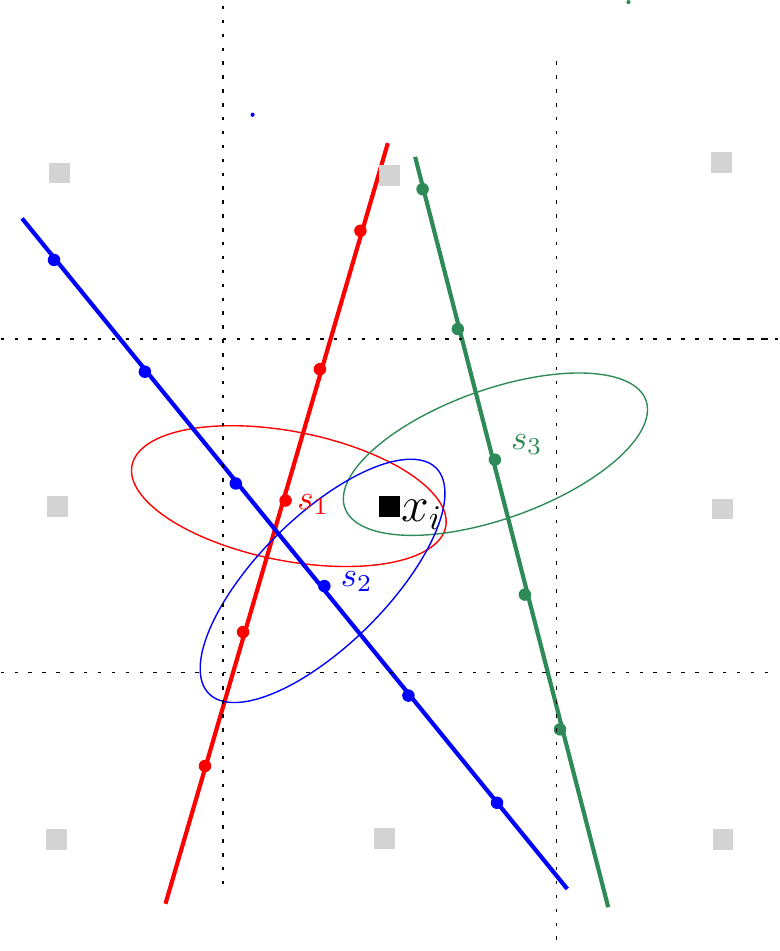}
	\caption{\shortcaption{Sample Selection Process}. Three rays pass through the voxel centered at $\bf{x}_i$. A sample $s_q$ is considered as relevant to $\bf{x}_i$ if it falls within the region of action of a sample defined by an ellipsoid (\cf \eqref{equ:ellipsoidSelection}). Only one sample per ray is taken into account.}
	\label{fig:sample_selection}
\end{center}
\end{figure}
\autoref{fig:sample_selection} shows the sample selection process exemplarily.
Based on \eqref{equ:ellipsoidSelection}, multiple samples of each scan line could be selected depending on the proximity of a ray to the target position ${\bf x}_i$.
As the main interest for the reconstruction is to collect complementary information, only the nearest sample of each ray is retained in $S_i$.

\subsection{Computational Aspects of CS}
\label{sec:computational}
After the samples have been selected, and a model (either tensor or discretized spherical function) has been computed per voxel, the 3D reconstruction process is finished, and the result is a 3D lattice $X$ of models. In this section we discuss the ``computational" aspect of CS, where we will make use of algorithms to extract the relevant information from the lattice both on-demand and according to the application. Within the scope of this paper we restrict the algorithms to the extraction of the information needed for visualization.

\subsubsection{Tensor visualization}
From the Tensor-CS reconstruction, it is possible to extract \emph{directional images}, \ie images of the reflected ultrasound signal from a given direction. The directional intensity $I_{ij}$ for a voxel $i$ from direction $\dir$, with $I : \mathbb{R}^{3 \times 3} \rightarrow \real$, can be estimated from the tensors with the second order approximation:
\begin{equation}
\hat{I_{ij}} = {\bf d_j}^\top T_i{\bf d_j}.
\label{equ:tensorProjection}
\end{equation}
Thereby, \eqref{equ:tensorProjection} allows us to interactively visualize the image as if taken from an arbitrary viewpoint, even if not previously imaged. Such interactive direction-dependent visualization can serve as a basis for realistic simulations for training.

Secondly, from the Tensor-CS we can also retrieve a variety of \emph{3D scalar representations} suitable for simple 3D renderings.
The first option is to compute the absolute value of the {\it tensor traces} and use them as the intensities. In this way the intensity at voxel $i$ becomes
$I_i=\textrm{abs}(\textrm{trace}(\tensor_i))$.
The tensor trace amounts to display the accumulated signal intensity from the three major components (eigenvectors) of the tensor. A second possibility is to extract the intensity from the tensor's dominant orientation corresponding to the largest eigenvalue of the tensor $I_i = max_{\bf y}\, {\bf y}^\top T_i {\bf y},\, \left|y\right|=1$.

\subsubsection{Visualization of discretized spherical functions}
In the case of the Spherical-CS, and in particular for the geodesic grids, we studied the visualization of two quantities: the mean and the maximum of the intensity values stored in the different cells of the geodesic grids. 
\begin{eqnarray}
I_i=\textrm{mean}(\tilde \function(k)) &  \forall k\in \{1, \ldots, N_{\rm cells}\},\\
\label{equ:spherical_median}
I_i=\textrm{max}(\tilde \function(k)) &  \forall k\in \{1, \ldots, N_{\rm cells}\}.
\label{equ:spherical_max}
\end{eqnarray}
These values are suitable for showing respectively, the most representative intensity in each voxel and the intensity from the direction with highest response. Examples for these visualizations can be seen in \autoref{subsec:qualitative-comparison}. These choices are motivated to maximize the speed of the visualization task. It is worth noting though that the representation is open to more complex reconstruction approaches.

\section{Experimental Validation}
\label{sec:Experimens}
To evaluate the different reconstruction capabilities of Computational Sonography we present next a series of experiments on both phantom and in-vivo datasets~(\autoref{subsec:datasets}).
First, we perform a quantitative evaluation of the CS representation quality in terms of information preservation and in comparison to classical compounding~(\autoref{subsec:quality-evaluation}).
We then investigate the robustness of the representation within a voxel to increasing variations of the intensity values, or to an increasing coverage of the acquisition directions~(\autoref{subsec:coverageErrorRelationship}).
We also visually illustrate the direct use of the CS representation to create scalar 3D volumes (\autoref{subsec:qualitative-comparison}). 
Finally, we present exemplary uses of the additional information present in CS volumes in \autoref{subsec:free_view_evaluation}.
The evaluation is based on a comparison of the following spatial compounding methods:
\begin{itemize}
	\item Mean compounding.
	\item Tensor-CS.
	\item Spherical-CS with latitude-longitude grid.  
	\item Spherical-CS with spherical Fibonacci grid. 
\end{itemize}
For all the Spherical-CS reconstructions we used the mean of sample intensities in each cell \eqref{eq:recofirst}.
For the Tensor-CS one constraint not enforced so far is the expected positive-definiteness of the tensors ($\dir^\top T_i\dir>0$), arising from the the fact that intensities are by nature positive.

\subsection{In-vitro and In-vivo Data Acquisition}\label{subsec:datasets}
\begin{table*}
	\footnotesize
	\centering
	\begin{tabular}{l | ccccccccccccccc|c}
		\backslashbox{Type}{Seq. No.}& 1 & 2 & 3 & 4 & 5 & 6 & 7 & 8 & 9 & 10 & 11 & 12 & 13 & 14 & 15 & Total\\
		\hline
		Vessel phantom & 125  & 217  & 556 &  633 &  725 &  1166 &  149  &  269  & 100  & 561 &     592     &   1000 & & & &6093 \\
		Carotid &160&    50   &292   &336   &437   &479   &224   &379    &25   &197   &209   &187   &204 & &  &3179 \\
		Femoral &89 &  377   & 312    & 78  & 209   & 188   & 306    & 81    & 31   &249    &91   &232   &204    &90   &187 & 2724\\
	\end{tabular}
	\caption{\shortcaption{Dataset description} including number of frames per sequence. The dataset consists of three types of image sequences including a vessel phantom as well as carotid and femoral in-vivo sequences. For every type we recorded several sequences, each with a varying coverage of the orientations.}
	\label{tab:imageNumbers}
\end{table*}

We acquired a number of phantom and in-vivo datasets using a  3D-Ultrasound system consisting of an Ultrasonix SonixMDP machine with a linear transducer and a NDI Polaris Vicra optical tracking system.
Every sweep is composed of several B-mode images acquired in a sequence while smoothly moving the probe.
All acquired 3D-US sweeps differ in the number of images and in the span of directions covered.
We followed a protocol where we gradually varied from sweeps taken for a single direction (as commonly done in today's 3D US freehand protocols) to sweeps densely covering the sphere. 

\begin{figure}
	\centering
	\includegraphics[angle=270,clip=true,trim=1.3cm 0 1.5cm 0,width=0.21\textwidth]{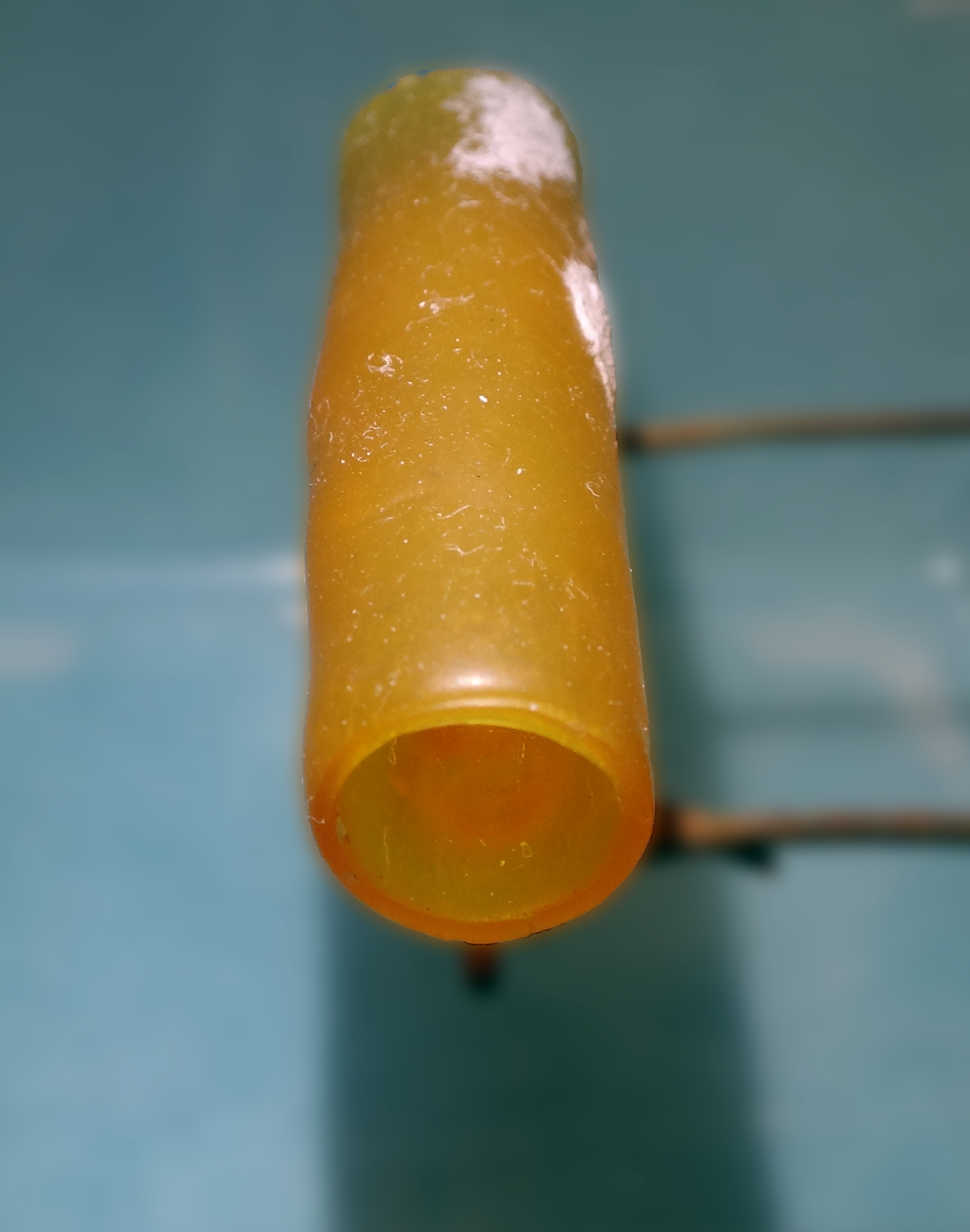}
		\caption{\shortcaption{Latex vessel phantom} used for static acquisitions.\label{fig:photo_vesselphantom}}
\end{figure}
In practice, we acquired 12 sweeps of a static latex vessel phantom (cf. \autoref{fig:photo_vesselphantom}) and 28 in-vivo sweeps, including 13 vascular scans (carotid arteries) and 15 scans in an osteological setting (femoral bones).
The number and size of the different acquisition sequences are detailed in \autoref{tab:imageNumbers} while the acquisition parameters are summarized in \autoref{tab:parameters}.
 
\begin{table}
	\footnotesize
	\centering
	\begin{tabular}{lc}
		Parameter & Value \\
	\hline
		US probe & L14-5/38 Linear \\
		Frequency & 6.6MHz \\
		Depth carotid & 4.0cm \\
		Depth femoral & 5.5cm \\
		Depth vessel phantom & 4.0cm \\
		Sample selection & $a=b=c=1mm$ \eqref{equ:ellipsoidSelection}\\
		Reconstruction resolution & 0.5mm spacing\\
	\end{tabular}
	\caption{\shortcaption{Acquisition parameters}}
	\label{tab:parameters}
\end{table}

\subsection{Representation Quality}\label{subsec:quality-evaluation}
Using all in-vitro and in-vivo datasets above we evaluate the representation capabilities of CS. First, we evaluate the representation quality in terms of how well a reconstructed volume preserves the information from the original ultrasound images,  
which is standard evaluation criterion for 3D reconstruction methods~\cite{solberg:usmb2007}. 
In particular, we focus on quantifying the preservation of the \emph{directional intensity}, \ie the intensity observed from different directions.  
To this end, and in contrast to the evaluation of classical compounding methods based on restricted scanning protocols (i.e. parallel images), our datasets contain information from various probe directions. 

To quantify the information preservation, we reproject the volumes at the positions and directions from where the original US images were acquired. For a volume reconstructed from a sequence including $N_\textrm{samp}$ samples $s_j$, we  measure the Mean Squared Error (MSE) between the input and the reprojected intensity values of every sample as follows:
\begin{equation}
e = \frac{1}{N_\textrm{samp}} \sum_{j=1}^{N_\textrm{samp}}(v_j-\hat v_j)^2,
\label{eq:rep-error}
\end{equation}
where $v_j$ is the intensity of sample $s_j$ and $\hat v_j$ is the intensity computed from the reprojection of the compounded volume at the position $\pos$ and direction $\dir$ of the original sample $s_j$. The error is averaged over all input samples.
For all experiments, 3D volumes were reconstructed with a 0.5mm spatial resolution. For the latitude-longitude grid a resolution of $30^{\circ}$ was used, whereas for the Fibonacci-geodesic reconstruction the grid was built with $512$ cells.
For scalar volumes, the reprojection simply corresponds to interpolation. 
In the case of tensor-CS, we reproject using \eqref{equ:tensorProjection}, whereas for spherical-CS we use the value of intensity stored in the associated grid cell.
Examples of the reprojections for each method are shown in image \autoref{fig:reprojections}. It can be seen that the 2D projections from computational-sonography volumes better preserve the information from the original B-mode images as CS avoids averaging information from different directions. Although the tensor-CS recovers appropriate intensity values for large portions of the image, the optimization fails in voxels where the information exceeds the low-complexity of the second order model or the symmetry. This is reflected as white regions in the images.



\def\reprofigwidth{0.22\linewidth}
\begin{figure}
	\centering
	\captionsetup[subfigure]{labelformat=empty}
	\setlength{\tabcolsep}{2pt}
	\begin{tabular}{c*{4}{c}}
	        \rotatebox{90}{\centering {\small Femur}}&
		\subfloat[]{\includegraphics[width=\reprofigwidth]{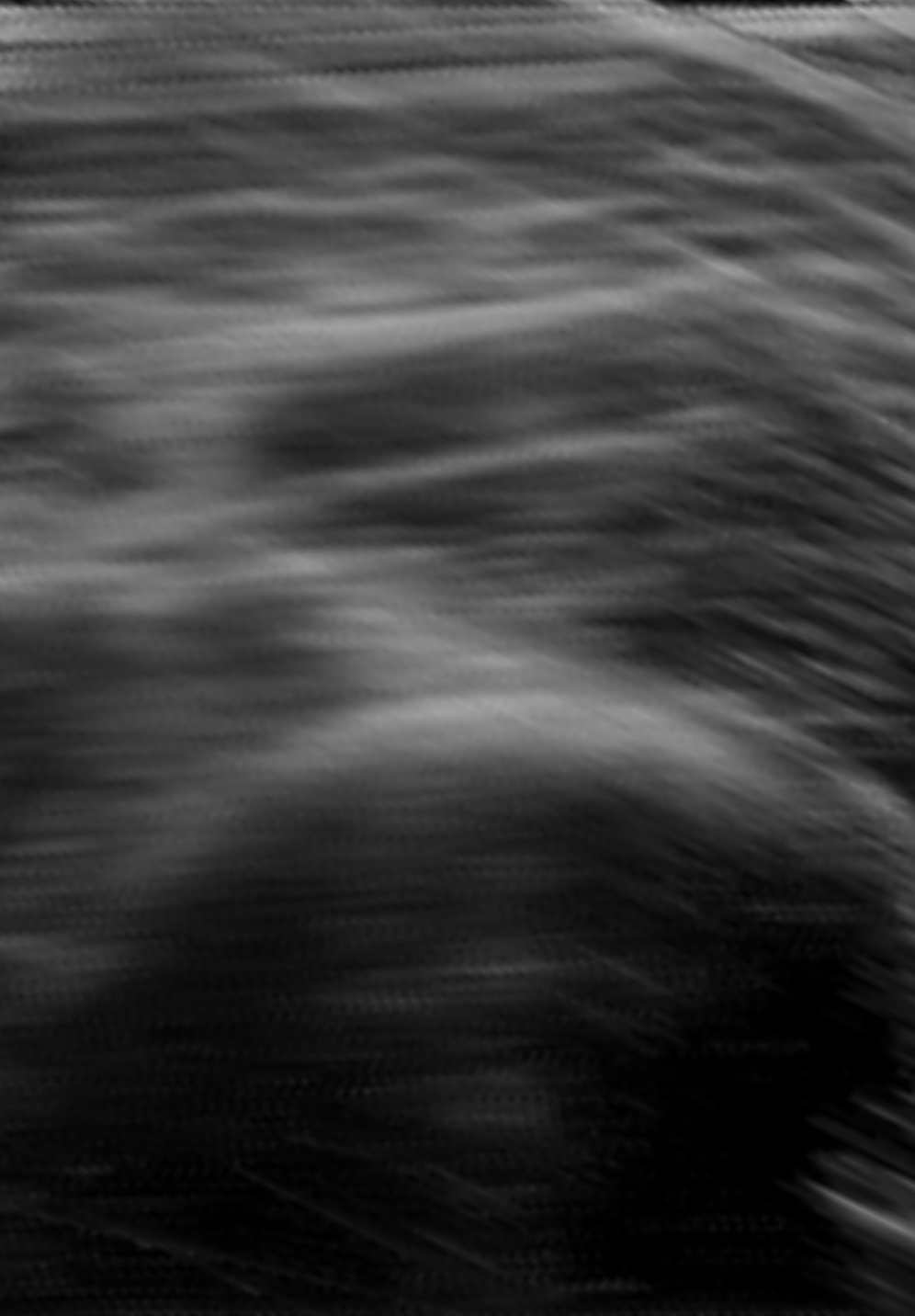}\label{fig:repro-femur-mean}}&
		\subfloat[]{\includegraphics[width=\reprofigwidth]{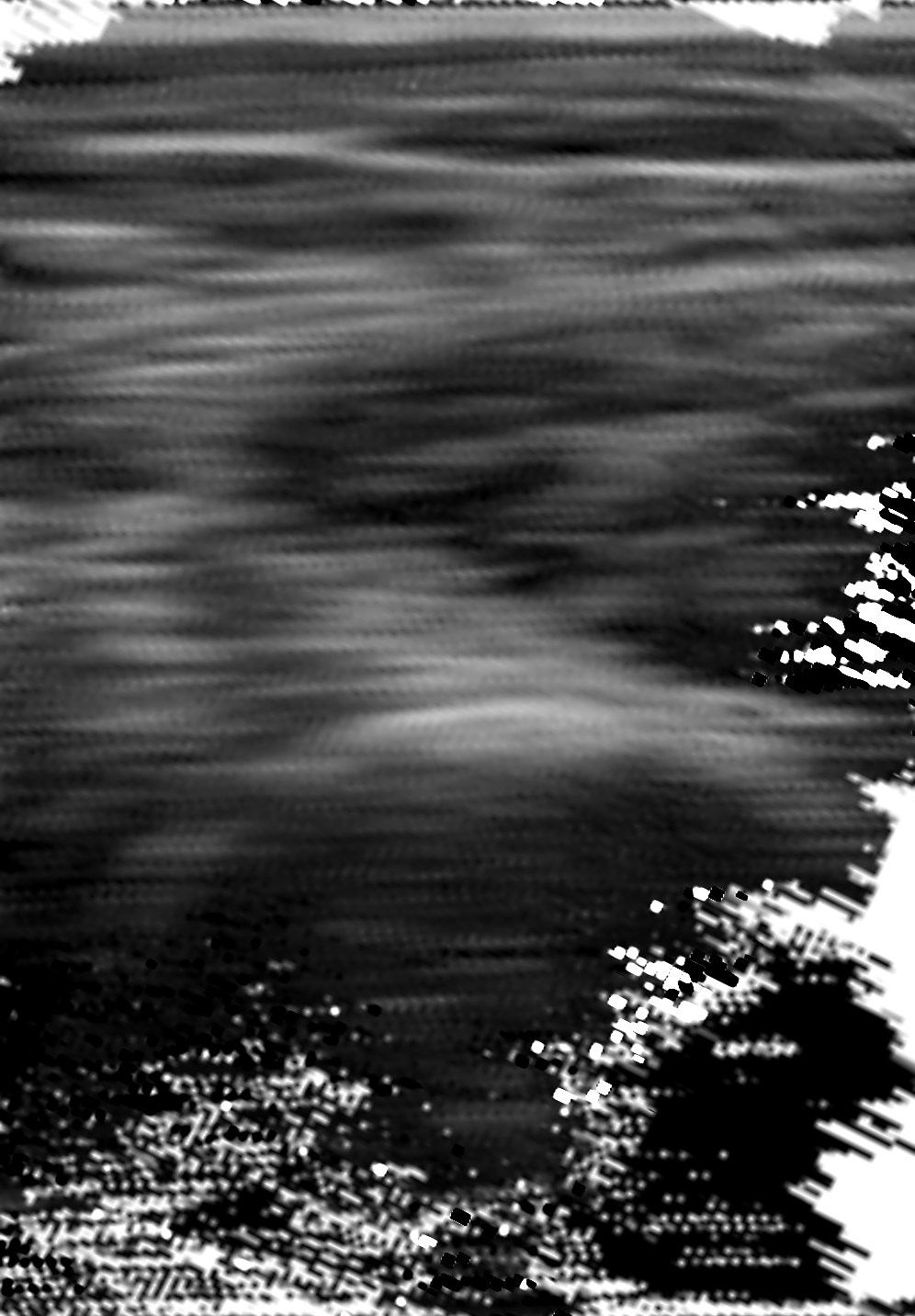}\label{fig:repro-femur-tensor}}&
		\subfloat[]{\includegraphics[width=\reprofigwidth]{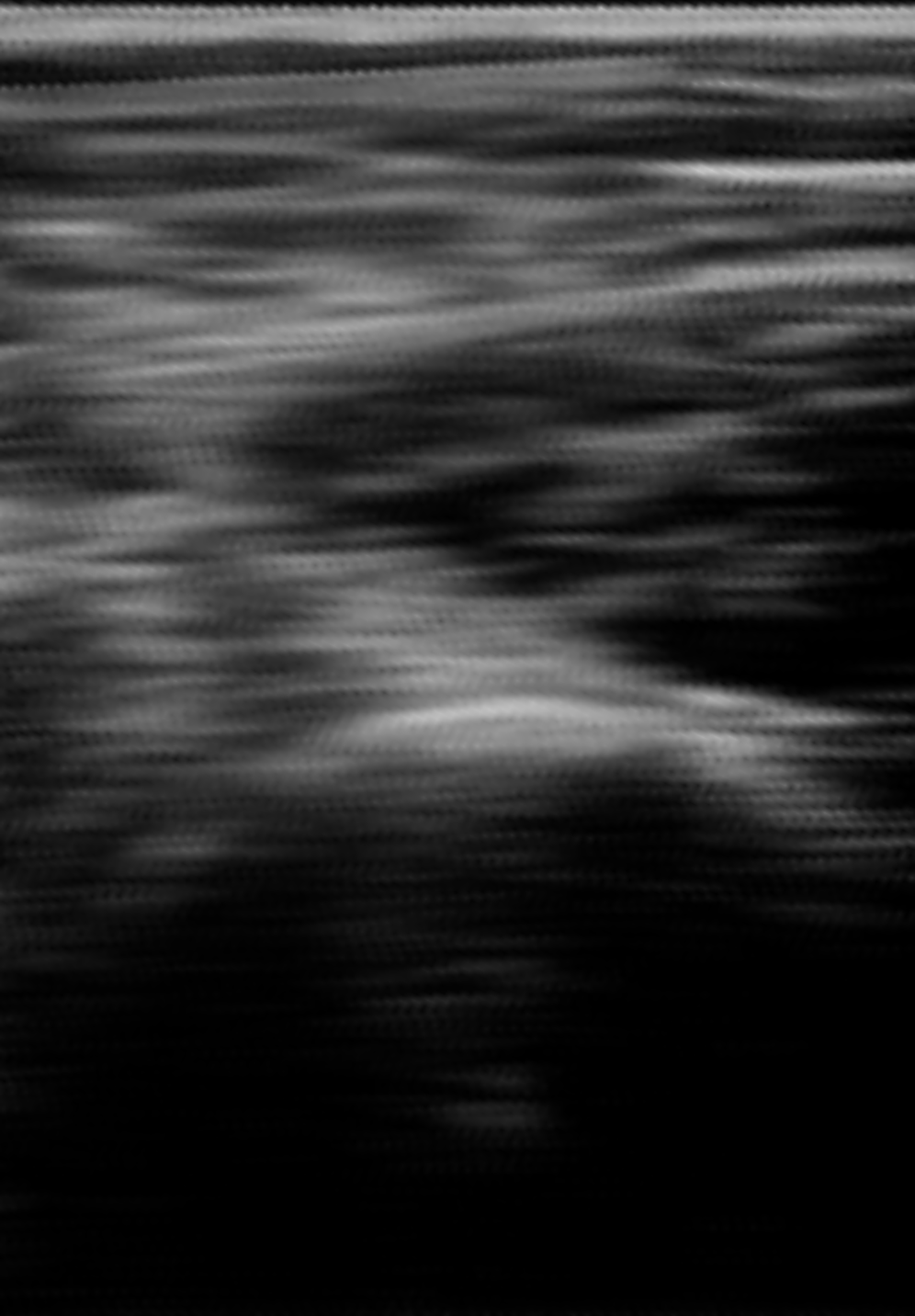}\label{fig:repro-femur-geodesic}}&
		\subfloat[]{\includegraphics[width=\reprofigwidth]{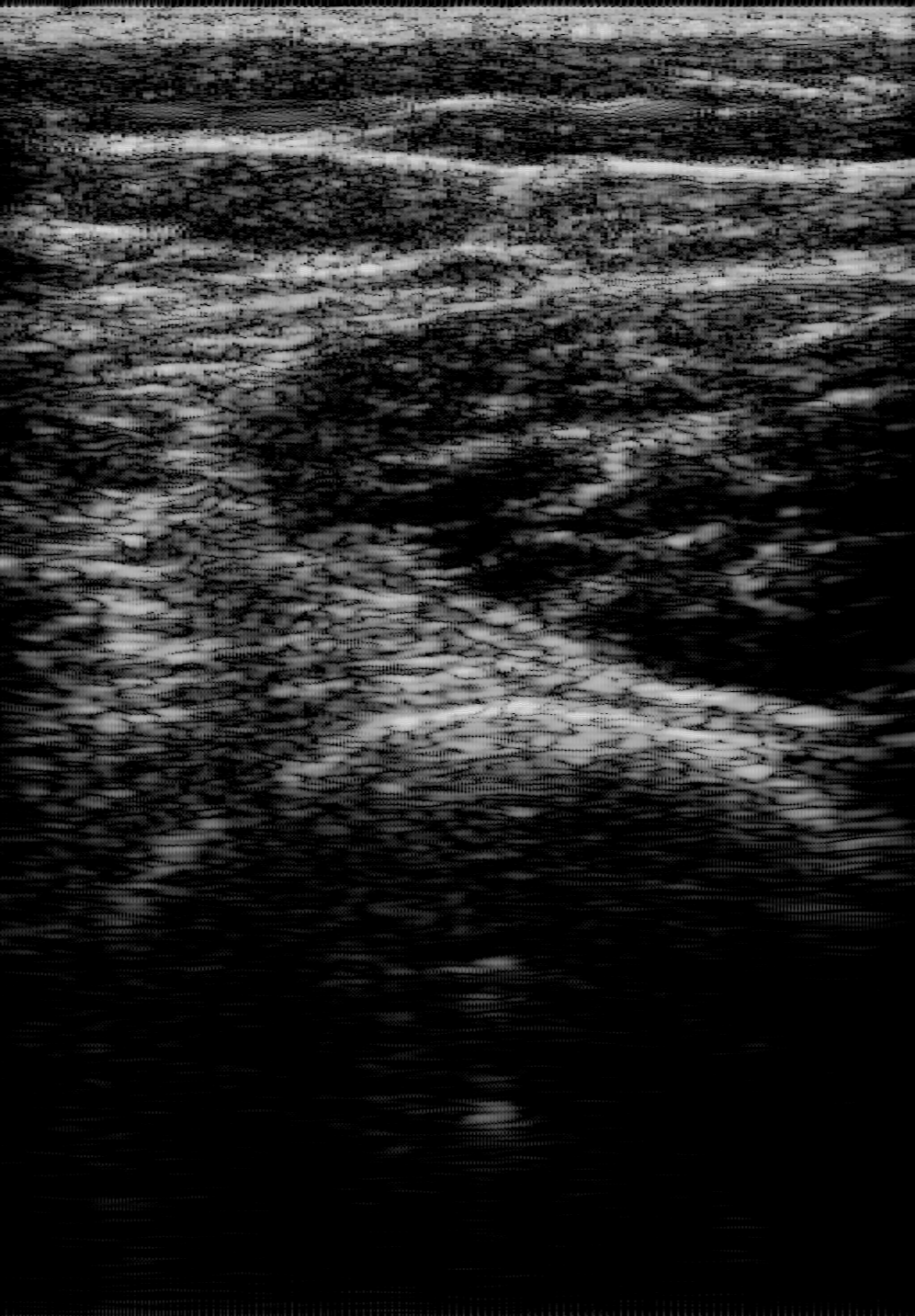}\label{fig:repro-femur-org}}
		\\[-20pt]
		\rotatebox{90}{\centering {\small Carotid}}&
		\subfloat[]{\includegraphics[width=\reprofigwidth]{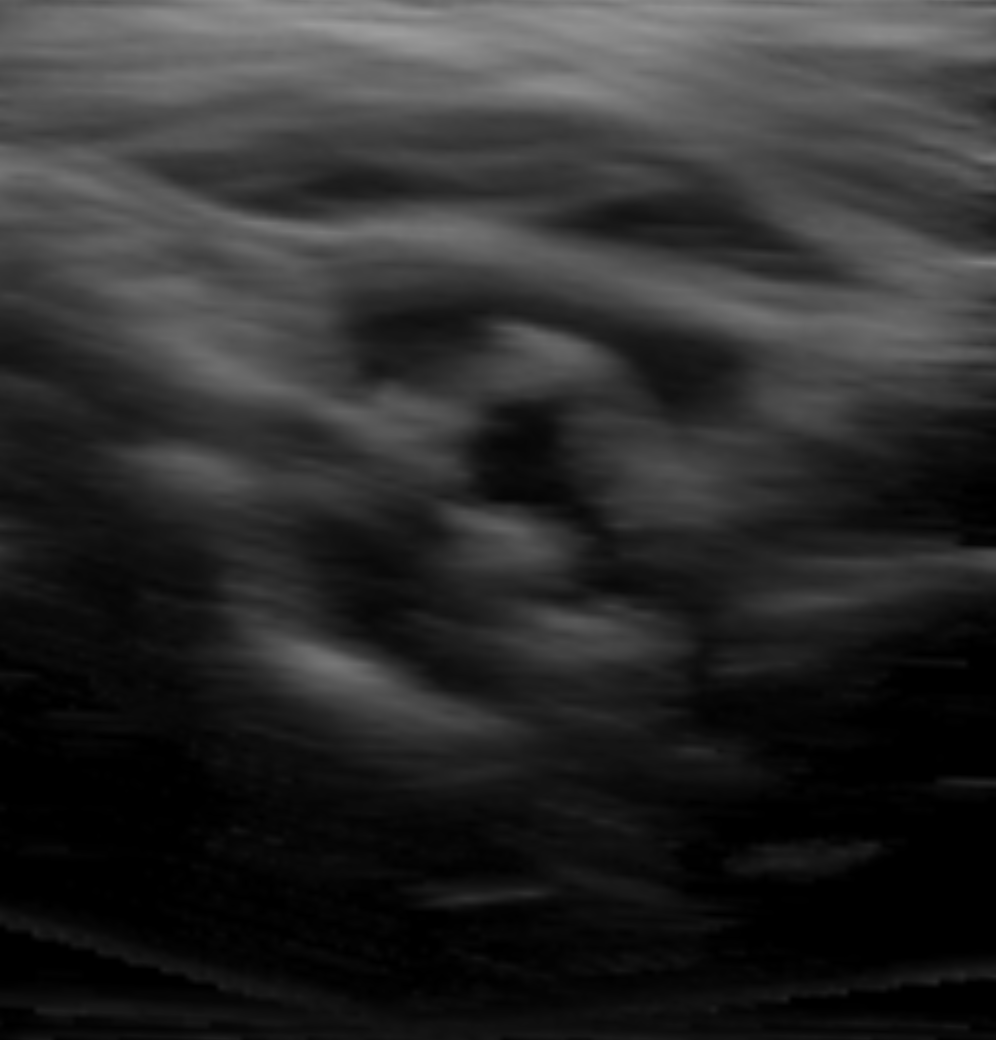}\label{fig:repro-carotid-mean}}&
		\subfloat[]{\includegraphics[width=\reprofigwidth]{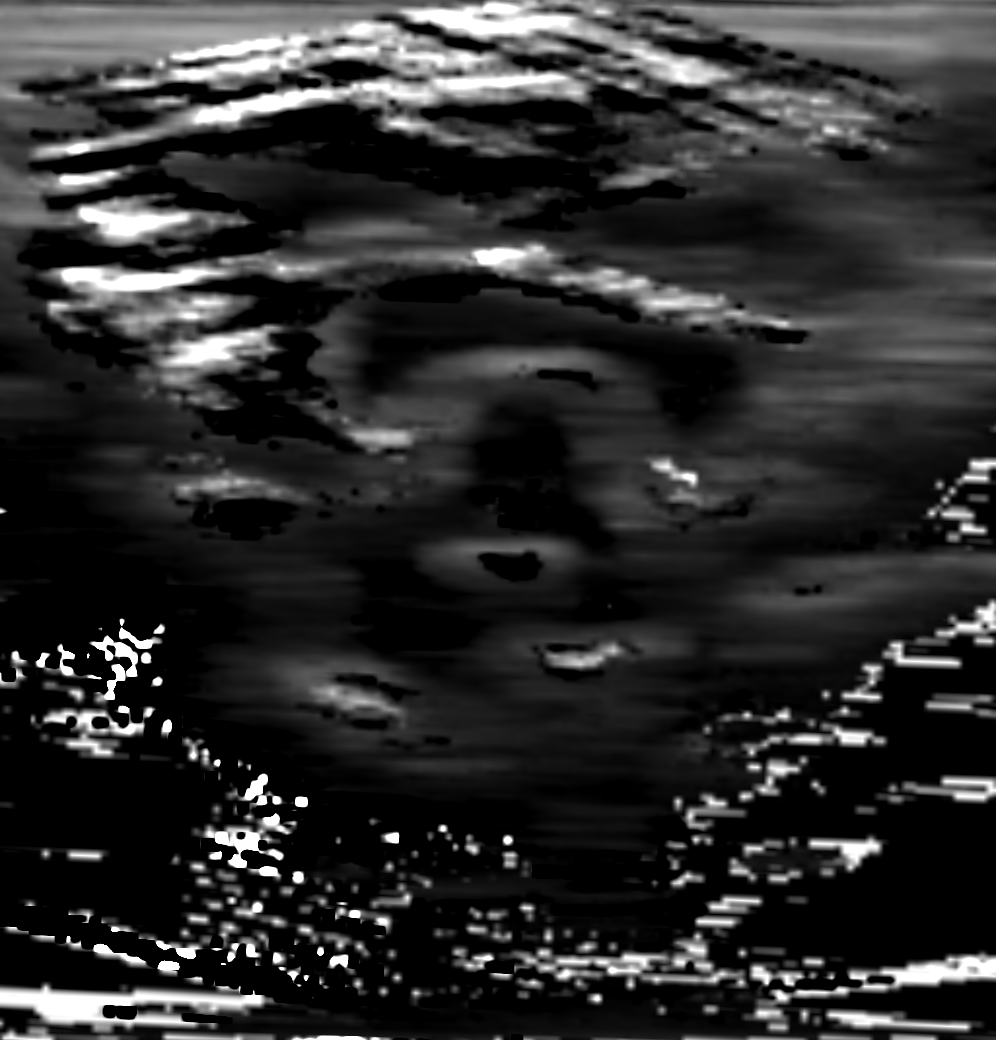}\label{fig:repro-carotid-tensor}}&
		\subfloat[]{\includegraphics[width=\reprofigwidth]{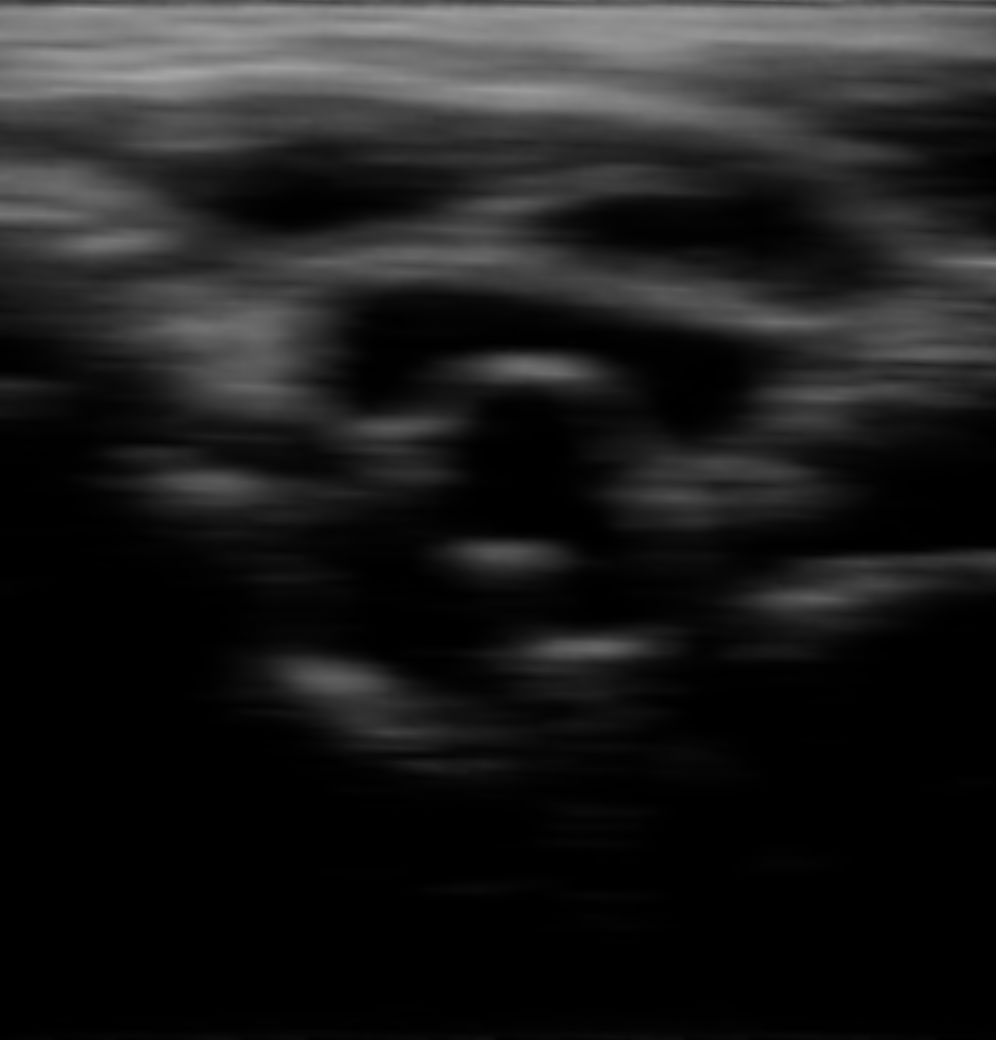}\label{fig:repro-carotid-geodesic}}&
		\subfloat[]{\includegraphics[width=\reprofigwidth]{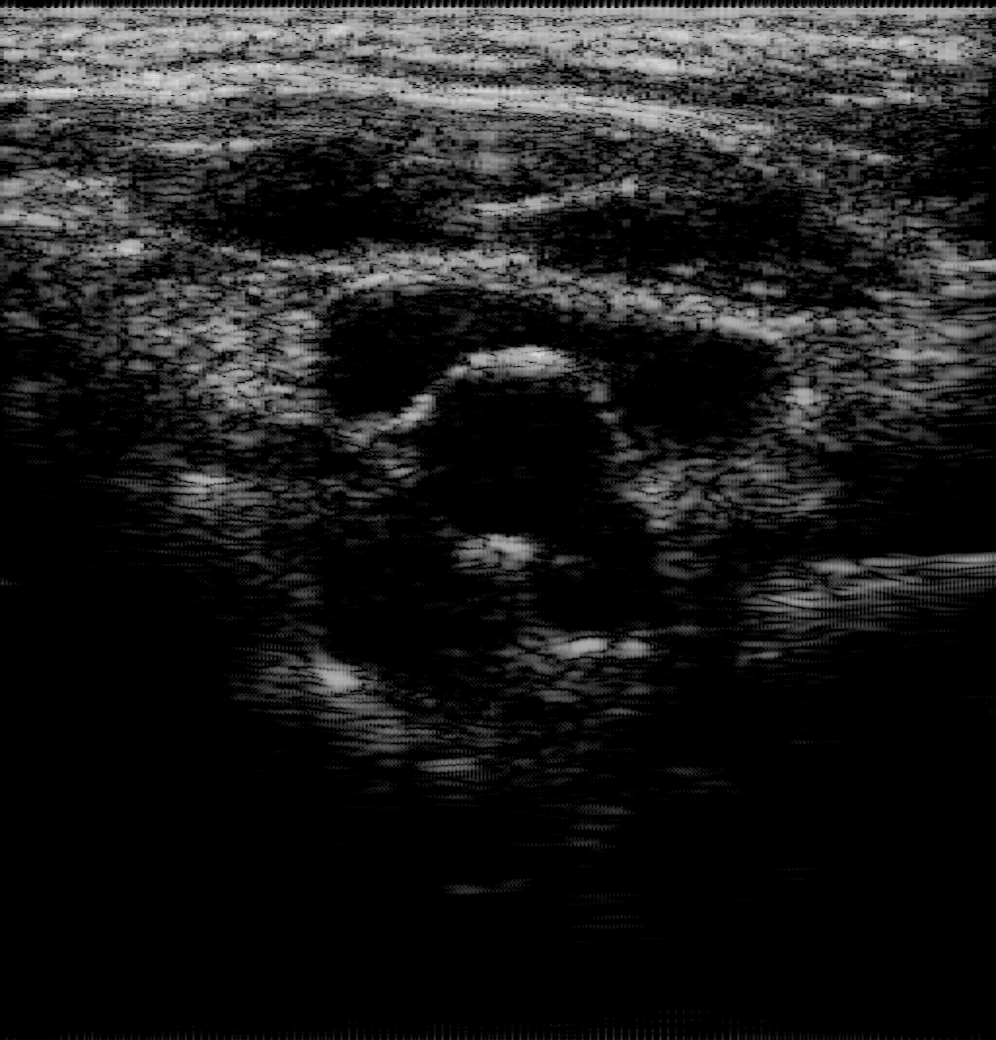}\label{fig:repro-carotid-org}}
		\\[-20pt]
		\rotatebox{90}{\small Vessel}&
		\subfloat[]{\includegraphics[width=\reprofigwidth]{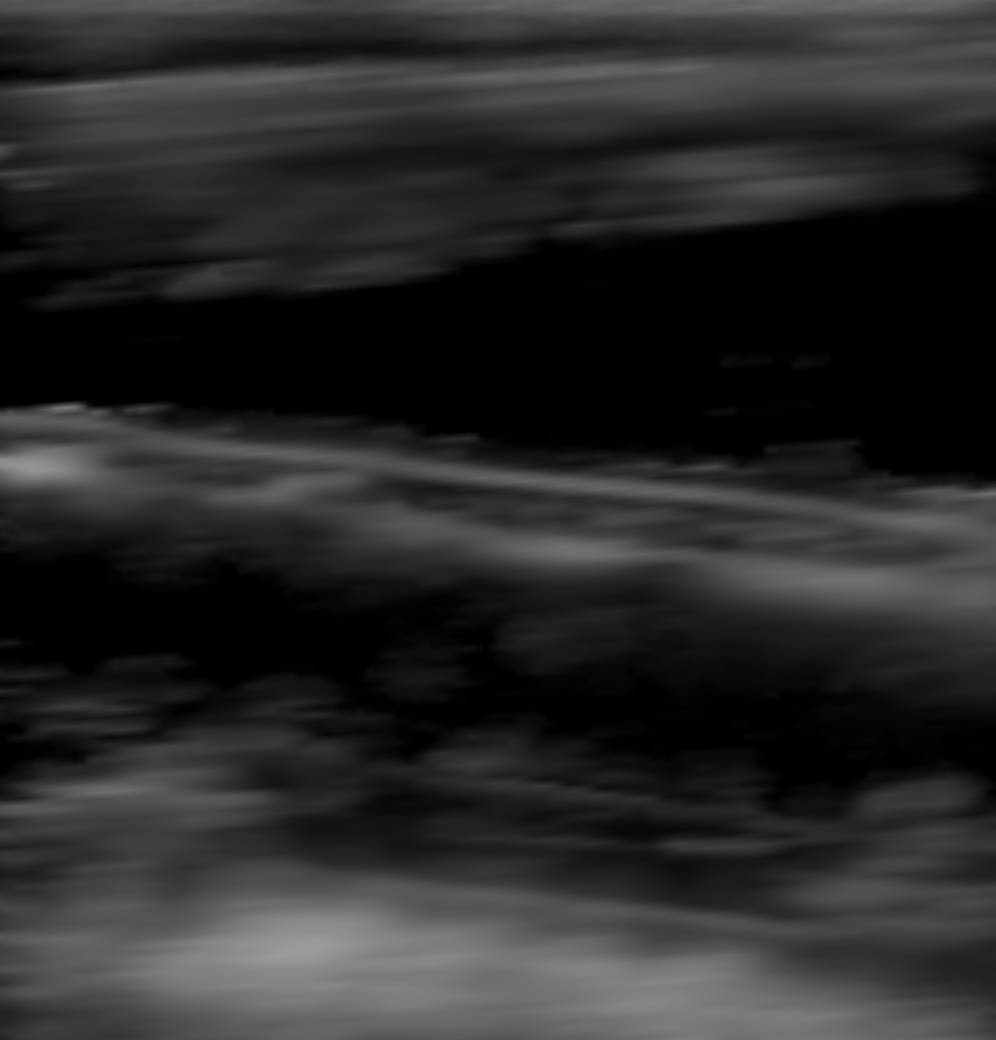}\label{fig:repro-vessel-mean}}&
		\subfloat[]{\includegraphics[width=\reprofigwidth]{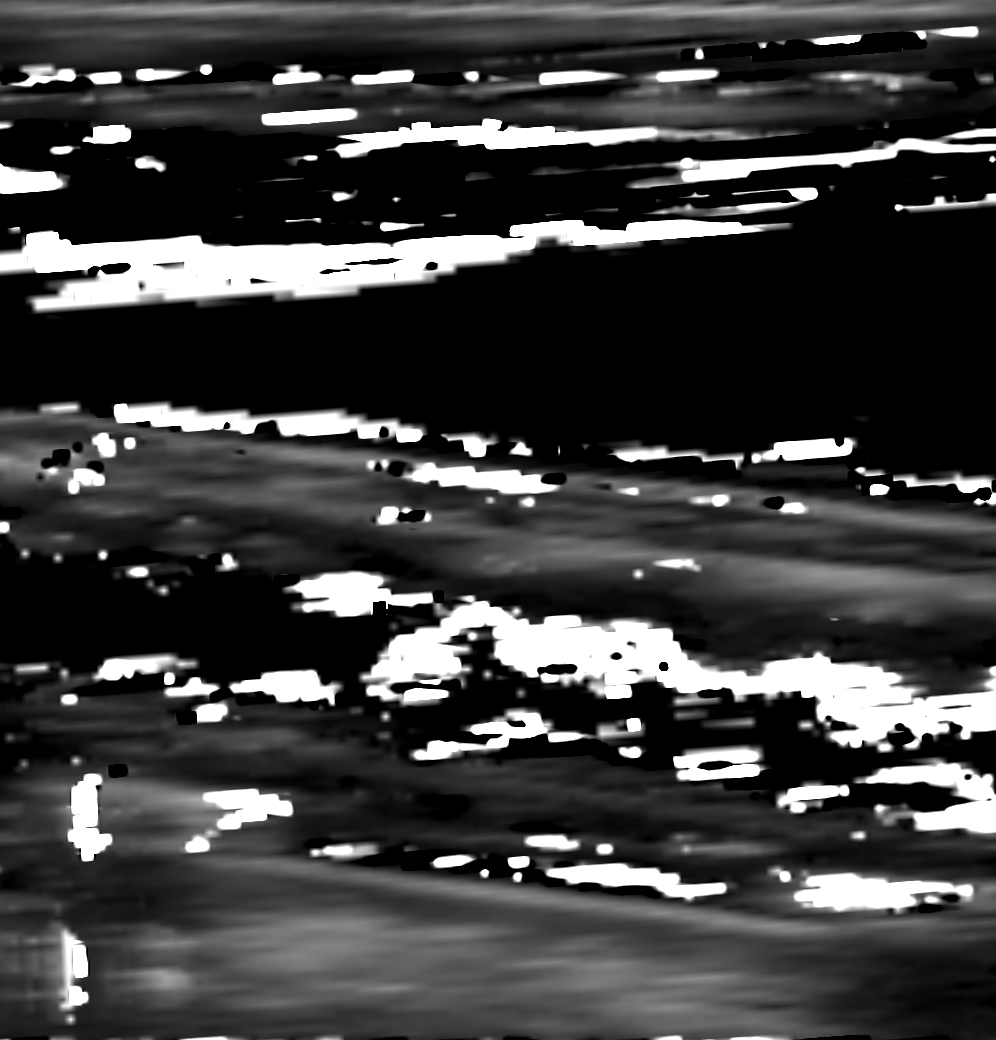}\label{fig:repro-vessel-tensor}}&
		\subfloat[]{\includegraphics[width=\reprofigwidth]{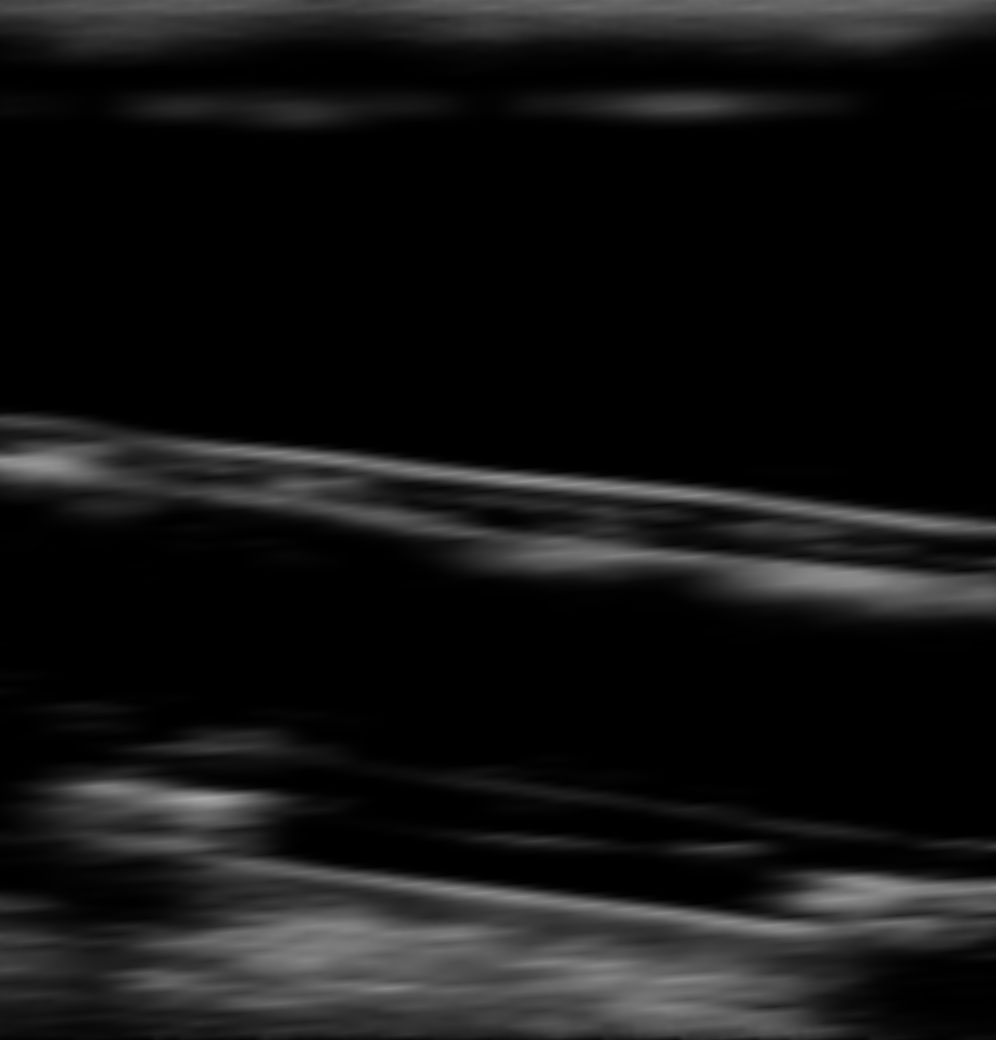}\label{fig:repro-vessel-geodesic}}&
		\subfloat[]{\includegraphics[width=\reprofigwidth]{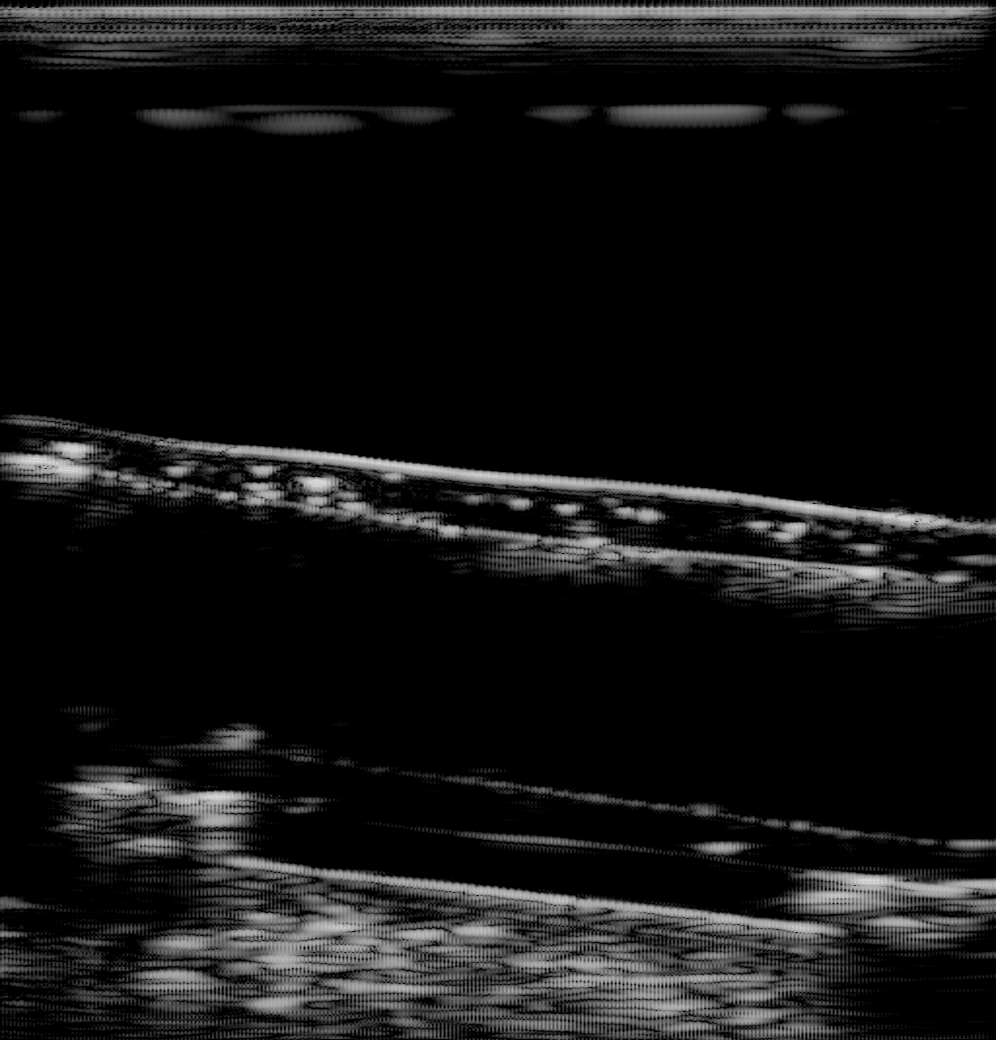}\label{fig:repro-vessel-org}}
		\\
		& 
		\begin{minipage}{\reprofigwidth} \centering {\small Mean compounding} \end{minipage}
		& 
		{\small Tensor-CS} & 
		\begin{minipage}{\reprofigwidth} \centering {\small Geodesic Spherical-CS} \end{minipage}&
		\begin{minipage}{\reprofigwidth} \centering {\small Original B-mode} \end{minipage}
	\end{tabular}
	\caption{\shortcaption{Reprojections}. 2D slices obtained by projecting the volumes reconstructed with  the three different methods in comparison to the original B-mode image.}
	\label{fig:reprojections}
\end{figure}

\begin{figure}
	\begin{center}
		\includegraphics[clip=true,trim=0 0 0 0,width=0.45\textwidth]{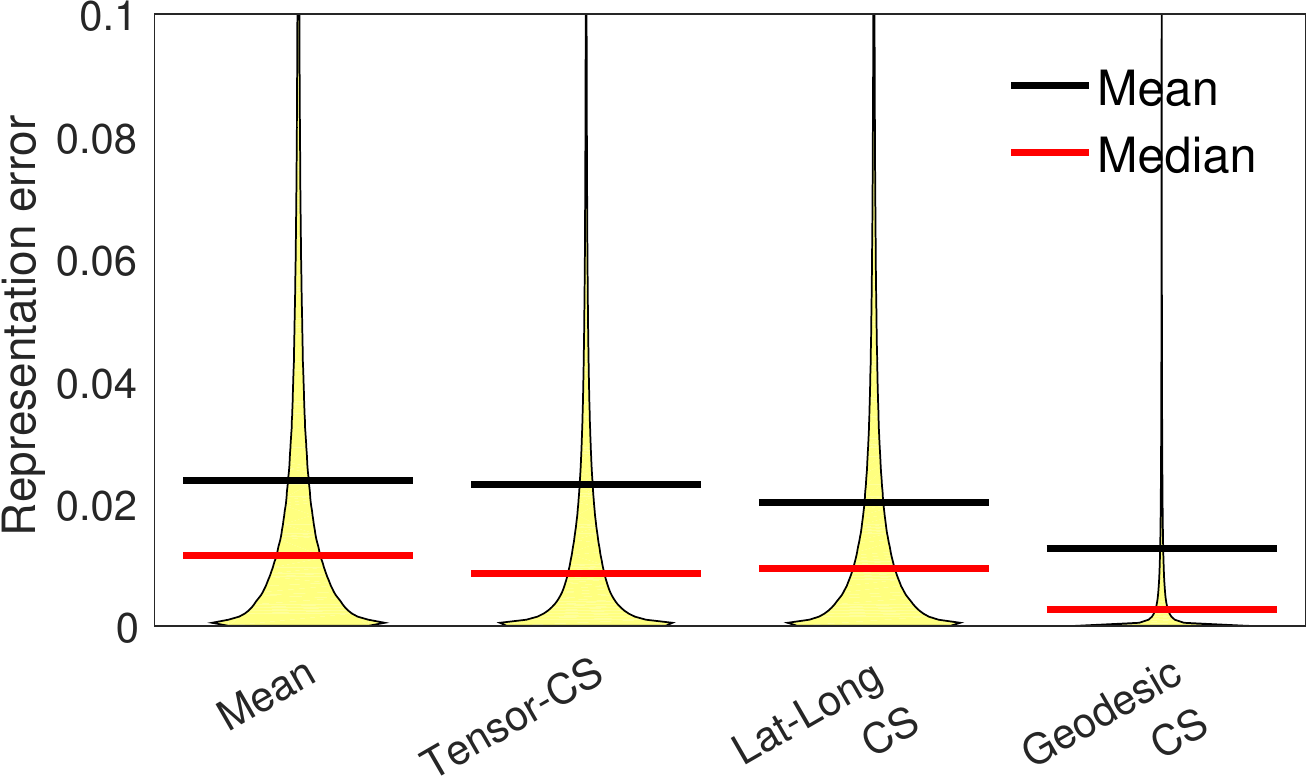}
	\end{center}
	\caption{
	\shortcaption{Violin plot of observed representation errors} in a carotid acquisition (Seq. 5, 437 frames).}
	\label{fig:error_violin}
\end{figure}

\begin{table}
	\footnotesize
	\centering
	\setlength{\tabcolsep}{2.7pt}
	\begin{tabular}{r c c c c c c }
		& Mean & Tensor-CS & Lat-Long-CS & Geodesic-CS\\ 
\hline 
Carotid & \meanStd{0.024}{0.034} & \meanStd{0.024}{0.092} & \meanStd{0.021}{0.031} & \meanStd{\textbf{0.014}}{0.028} \\ 
Femur & \meanStd{0.028}{0.039} & \meanStd{0.038}{0.157} & \meanStd{0.022}{0.034} &  \meanStd{\textbf{0.015}}{0.030} \\ 
In-vitro & \meanStd{0.025}{0.031} & \meanStd{0.061}{0.242} & \meanStd{0.020}{0.029} & \meanStd{\textbf{0.008}}{0.024} \\ 
\hline 
Overall & \meanStd{0.026}{0.036} & \meanStd{0.040}{0.170} & \meanStd{0.021}{0.032} & \meanStd{\textbf{0.013}}{0.028} \\ 
	\end{tabular}
	\caption{\shortcaption{Mean and SD of Squared Representation Errors.} For the in-vivo and in-vitro acquisitions, comparing representations of B-mode intensities, for voxels for which a tensor could be estimated.}\label{tab:overalltable_tensor}
\end{table}

We report in \autoref{tab:overalltable_tensor} the mean and standard deviation of the representation error in \eqref{eq:rep-error} for all volumes computed from the in-vivo and in-vitro acquisitions. For a fair comparison with the tensor-CS, we considered only the representation error of voxels for which a tensor could be estimated. 
The results indicate that representing the B-mode data by tensors leads to errors comparable to mean compounding, which is in accordance to the results in our prior work~\cite{hennersperger2015computational}. Spherical-CS based on Latitude-longitude grids does not substantially reduce the representation error, whereas geodesic-based spherical-CS results in the lowest average errors for all datasets.
\autoref{fig:error_violin} shows violin plots of the representation errors found in a carotid acquisition (Seq. 5, 437 frames, ROI dir.var.~\meanStd{0.0257}{0.0061}) and emphasizes the similarity of error distributions between the presented representations.


\subsection{Robustness to Intensity Variability and Spherical Coverage}
\label{subsec:coverageErrorRelationship}

In the following experiments, we investigate the robustness of the compared methods to two potential sources of inconsistency within the samples of a voxel: intensity variability and directional coverage.

\begin{figure}
	\centering
	\subfloat[]{
		\includegraphics[width=0.85\linewidth]{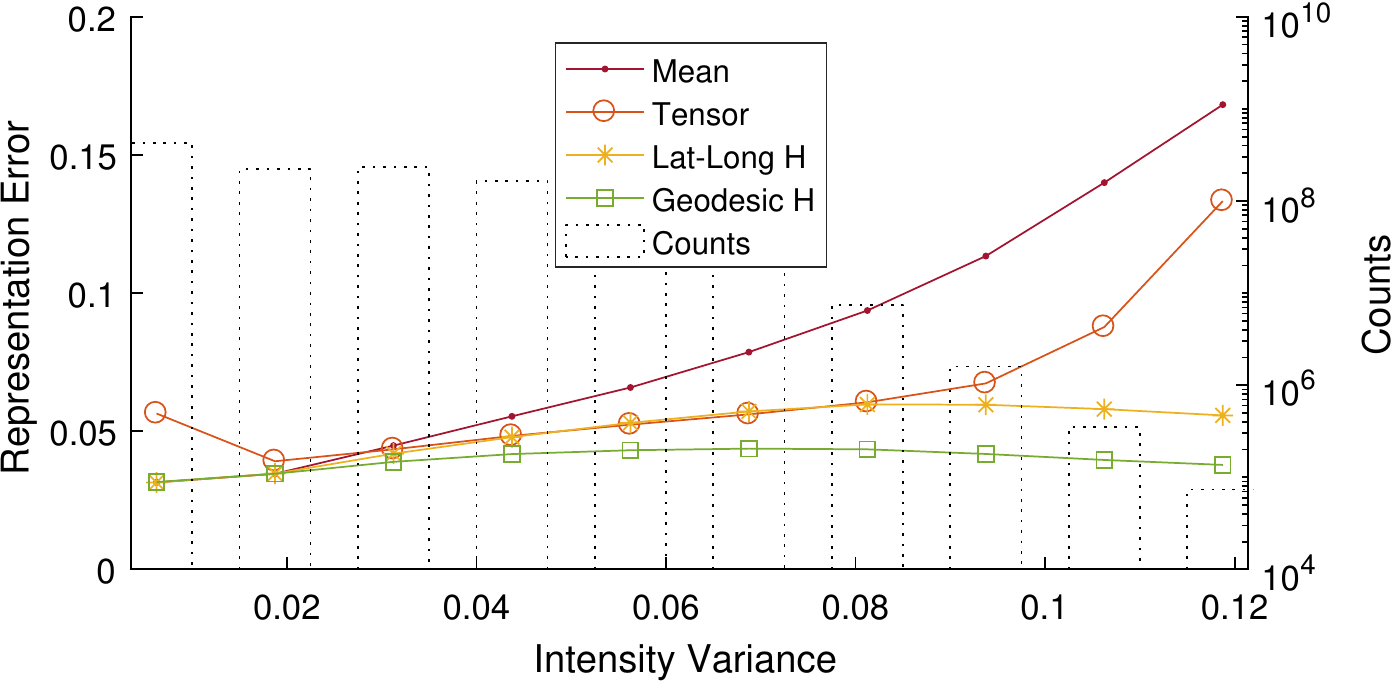}
		\label{fig:variance_intensity_invivo}
	}
	\\
	\subfloat[]{
		\includegraphics[width=0.85\linewidth]{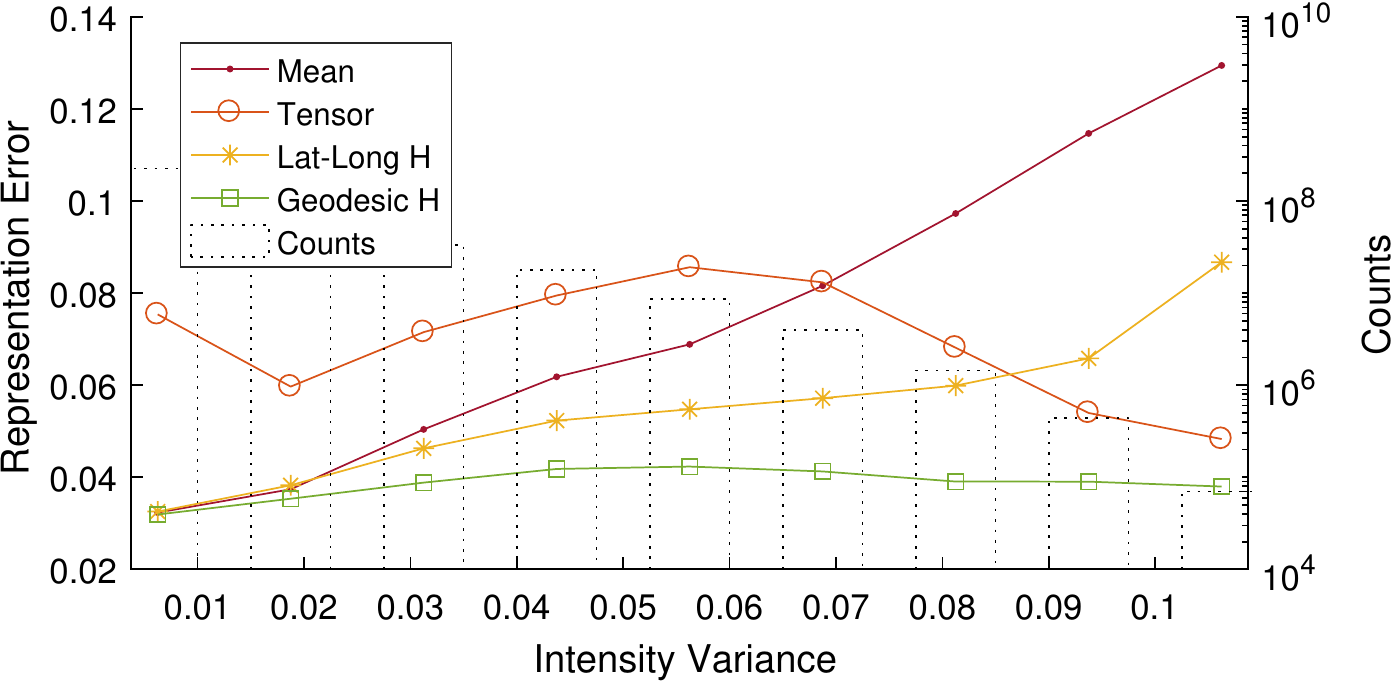}
		\label{fig:variance_intensity_phantom}
	}
	\\
	\subfloat[]{
		\includegraphics[width=0.85\linewidth]{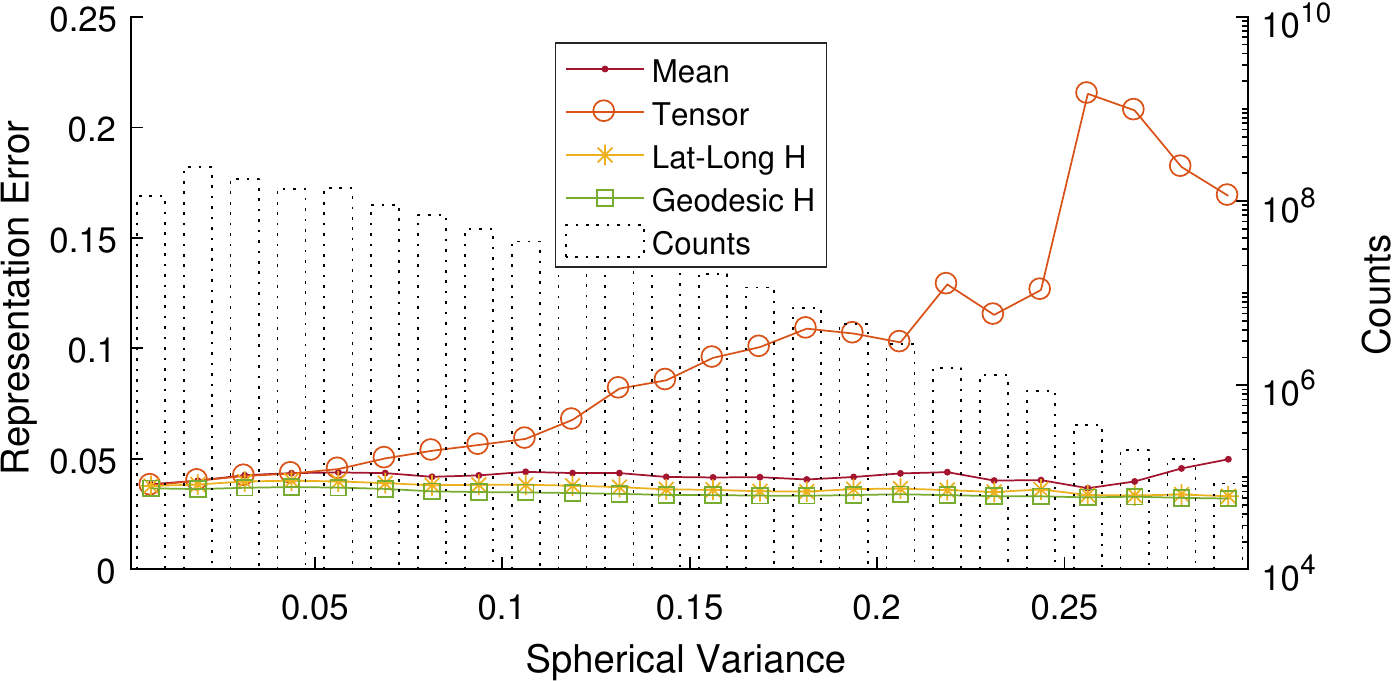}
		\label{fig:variance_direction_invivo}
	}
	\\
		\subfloat[]{
		\includegraphics[width=0.85\linewidth]{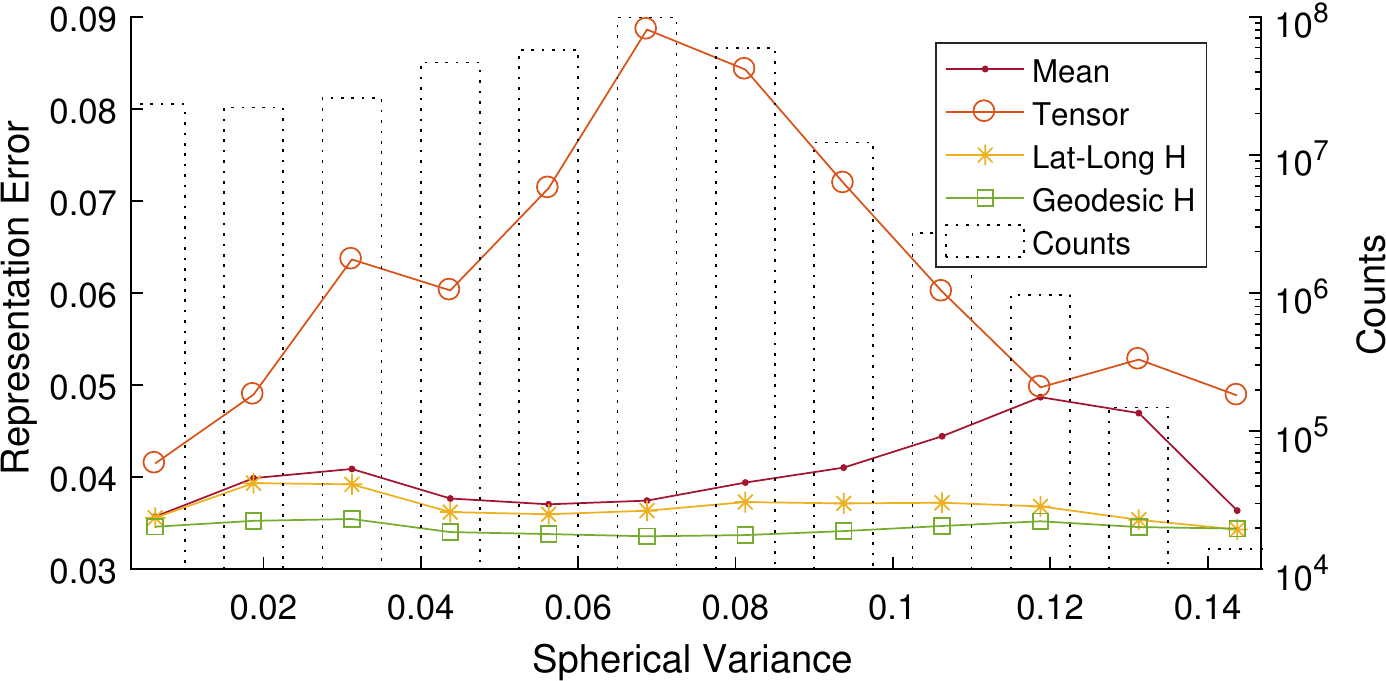}
		\label{fig:variance_direction_phantom}
	}
	\caption{\shortcaption{Robustness against intensity variability and directional coverage.}  
	\mysubref{a} \& \mysubref{b} Mean square representation error for increasing intensity-spans within each voxel, for the in-vivo sweeps in \mysubref{a} and the vessel phantom sweeps in \mysubref{b}. 
	\mysubref{c} \& \mysubref{d} Mean square representation error for an increasing directional coverage (angle-span) within each voxel.
	\mysubref{c} shows the errors for the in-vivo acquisitions, \mysubref{d} for the phantom sweeps.}
	\label{fig:coverage_error_Bmode}
\end{figure}

To evaluate the robustness against the {\it intensity-span}, we calculate the representation error for an increasing variance of intensity values associated to a voxel.
Given intensities normalized to $[0,1]$, we computed the variance of the intensities for each voxel $\sigma_{\rm int}(i) := {\rm var}(\{v_j\})\,\, \forall s_j \in S_i$ , and used a histogram to group voxels with similar variance. 
Figures \ref{fig:variance_intensity_invivo} and \ref{fig:variance_intensity_phantom} show the representation error for the different quantized levels of intensity-variations. It can be clearly seen that all Computational Sonography methods results in significantly lower errors, especially  for larger values of intensity variance. Moreover for Spherical-CS the error is stable.

We also analyze robustness to the {\it spherical-span}, that is, how the variance in the span of directions covered by the acquisition affects the representation error. 
We repeat the procedure above, but quantizing the spherical variance $\sigma_{\rm sph}(i)=2(1-\overline{R}(i))$, as defined by Mardia and Jupp \cite{mardia2009directional}, where
$\overline{R}(i) = || \frac{1}{|S_i|} \sum_{j=1}^{|S_i|} \dir || $ is the so called mean resultant length, which depends on the clustering of the vectors $\dir$ around the the mean direction. Note that $\sigma_{\rm sph}(i)$ is in $[0, 2]$, corresponding to acquisitions going from a single direction to uniformly distributed over the sphere.
We then estimate the representation error for the different levels of spherical variance.
Figures \ref{fig:variance_direction_invivo} and \ref{fig:variance_direction_phantom} shows how an increasing variance of directions affects the tensor representation the most, which is explained by the fact that the symmetry assumption is more likely broken in voxels compounding samples with high spherical variance. 
For the mean-compounding and spherical-CS the increase of spherical spans does not seem to have a noticeable impact on the reconstruction error. 

The two graphs provide an experimental evidence to the fact that classical compounding approaches result in reconstruction errors when applied to unconstrained free-hand acquisitions. The reason for this is not directly the increased spherical variance but rather the ultrasound physics causing the same structure to have large intensity variations in different views.
With regard to the two different CS representations, 
the tensor-CS suffers from a limited 2nd order model complexity, which reflects in a poor performance under large intensity variations 
On the other side, the strongest advantage of spherical-CS comes for acquisitions with heterogeneous ultrasound information (i.e. strong intensity variations) or high spherical variance. 
In fact, the error of spherical-CS reconstructions decreases for larger spherical variances, better achieving the goal of preserving the directional information, although at the cost of increased model complexity.
These results further confirm computational sonography as a viable alternative to traditional reconstruction methods, allowing for high quality 3D free-hand reconstructions from arbitrary scanning trajectories instead of restricted scanning protocols.




\subsection{Qualitative comparison}\label{subsec:qualitative-comparison}
\begin{figure}
	\begin{center}
		\includegraphics[width=0.35\textwidth,clip,trim=0 0 0 0]{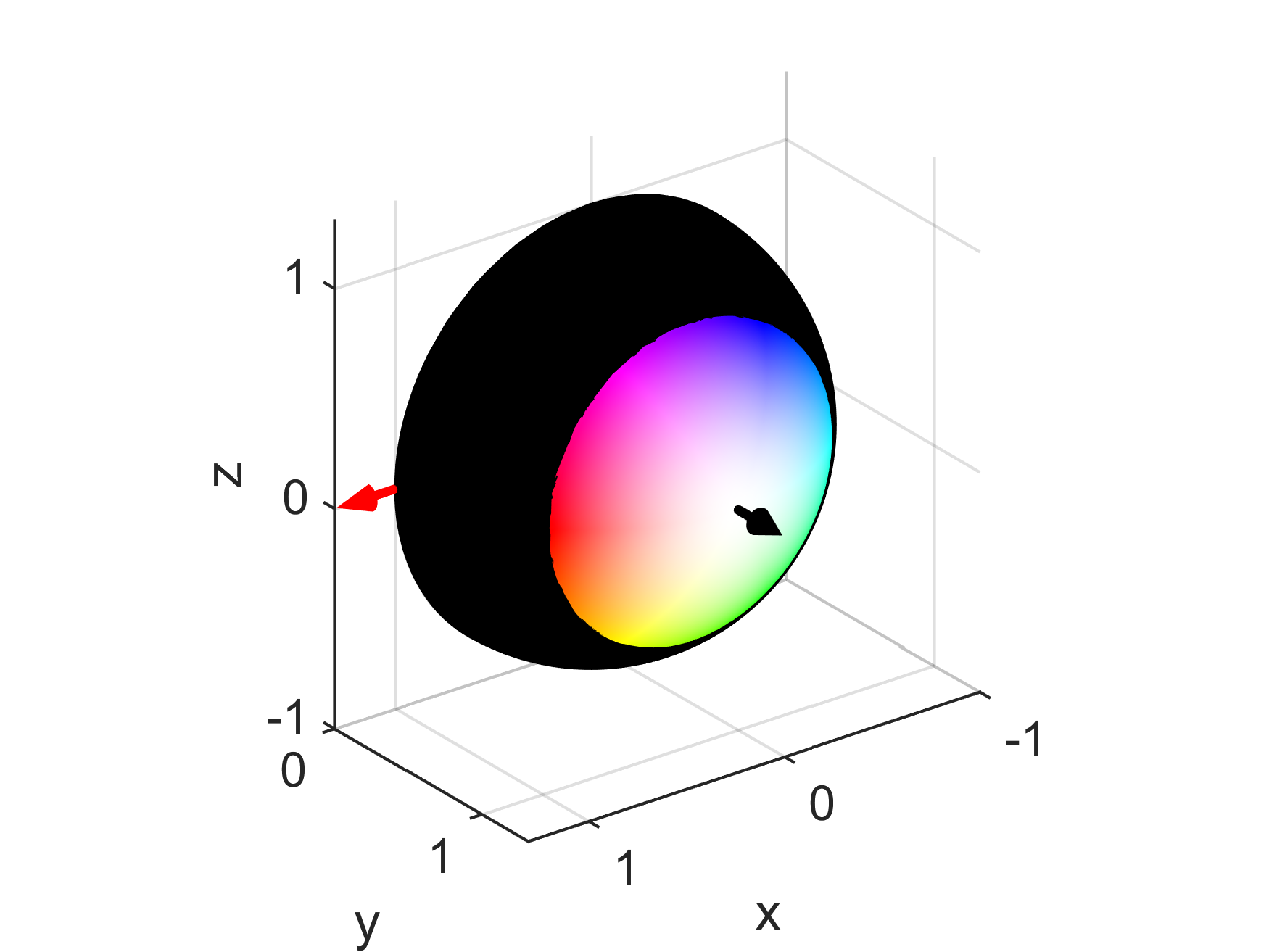}
	\end{center}
	\caption{\shortcaption{Direction color-map} w.r.t. the per-dataset main direction (black) and the direction determining the orientation of the map (red).	\label{fig:direction-color-map}}
\end{figure}

Exemplary results of the different reconstruction methods are shown in Fig. \ref{fig:volumes-qualitative}. 
They include axial and lateral slices for each of the sequence types: femoral, carotid and vessel phantom. 
For tensor-CS, we show the absolute value of the tensor trace. 
For spherical-CS, we use two forms of visualization: first, the intensities reflecting the mean of the cell values excluding empty cells, and second, the color-coded direction of the cell with the highest value.
The color coding maps each vector in the volume to a unique color (\cf \autoref{fig:direction-color-map})\footnote{In practice, we first estimate the per-acquisition main direction by means of averaging the direction vectors in the volume.
The color-coding then shows deviations from the main direction with increasing saturation, the hue is determined by the angle on the plane perpendicular to the main direction.}
Because the map considers 3D direction vectors, similar colors in different views stand for similar maximum directions.

\def\thisfigwidth{0.2\linewidth}
\begin{figure*}
	\centering
	\captionsetup[subfigure]{labelformat=empty}
	\setlength{\tabcolsep}{2pt}
	\renewcommand{\arraystretch}{0}
	\begin{tabular}{ccccc}
		\rotatebox{90}{\centering Femur lat.}&
		\subfloat[]{\includegraphics[width=\thisfigwidth,clip=true,trim=14 80 62 68]{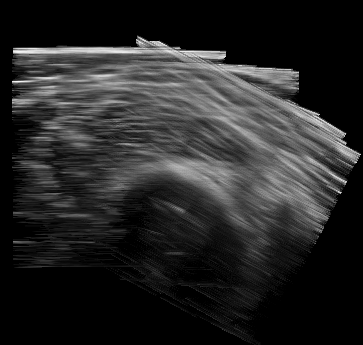}\label{fig:slices-femur-mean-lateral}}&
		\subfloat[]{\includegraphics[width=\thisfigwidth,clip=true,trim=14 80 62 68]{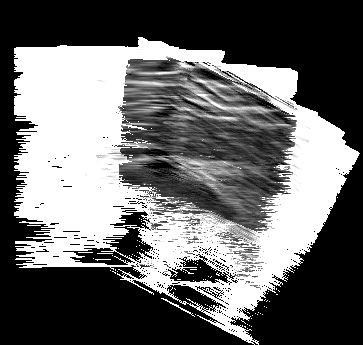}\label{fig:slices-femur-trace-lateral}}&
		\subfloat[]{\includegraphics[width=\thisfigwidth,clip=true,trim=14 80 62 68]{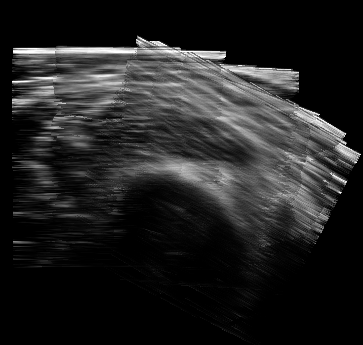}\label{fig:slices-femur-geomean-lateral}}&
		\subfloat[]{\includegraphics[width=\thisfigwidth,clip=true,trim=14 80 62 68]{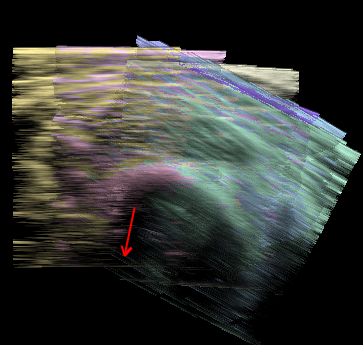}\label{fig:slices-femur-geomax-lateral}}
		\\[-2em]
		\rotatebox{90}{\centering Femur ax.}&
		\subfloat[]{\includegraphics[width=\thisfigwidth,clip=true,trim=72 190 110 86]{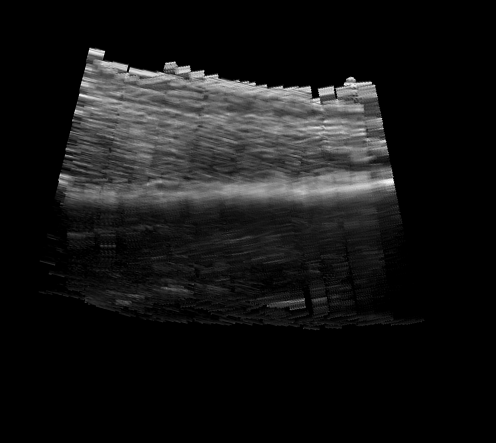}\label{fig:slices-femur-mean-axial}}&
		\subfloat[]{\includegraphics[width=\thisfigwidth,clip=true,trim=72 190 110 86]{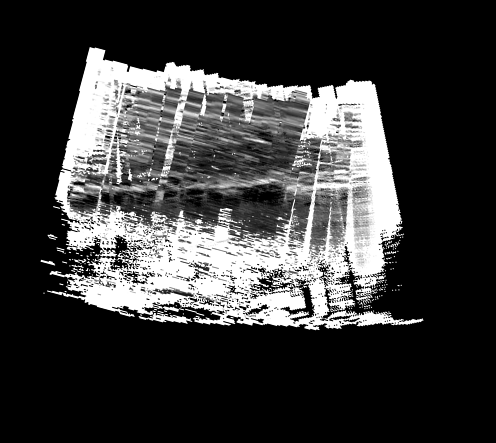}\label{fig:slices-femur-trace-axial}}&
		\subfloat[]{\includegraphics[width=\thisfigwidth,clip=true,trim=72 190 110 86]{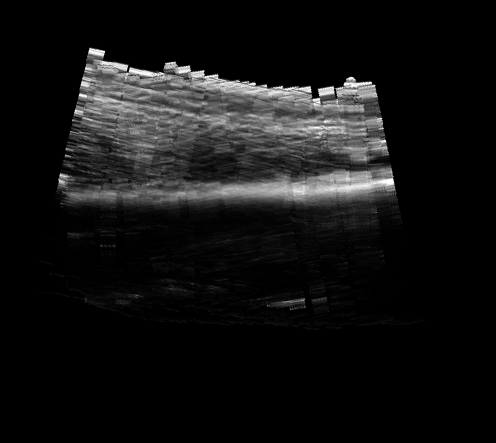}\label{fig:slices-femur-geomean-axial}}&
		\subfloat[]{\includegraphics[width=\thisfigwidth,clip=true,trim=72 190 110 86]{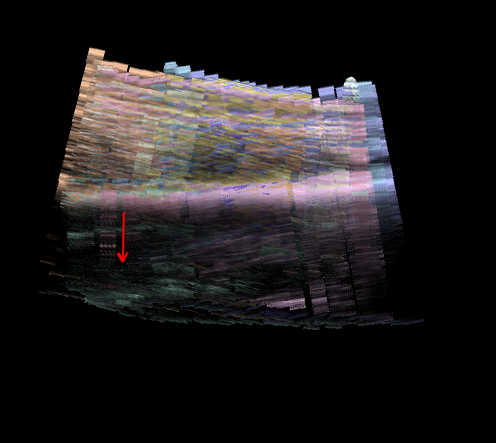}\label{fig:slices-femur-geomax-axial}}
		\\[-2em]
		\rotatebox{90}{\centering Carotid lat.}&
		\subfloat[]{\includegraphics[width=\thisfigwidth,clip=true,trim=39 80 23 56]{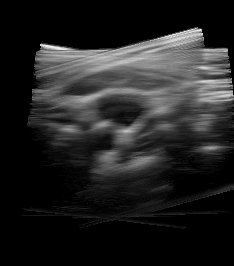}\label{fig:slices-vessel-mean-lateral}}&
		\subfloat[]{\includegraphics[width=\thisfigwidth,clip=true,trim=39 80 23 56]{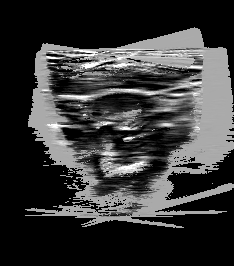}\label{fig:slices-vessel-trace-lateral}}&
		\subfloat[]{\includegraphics[width=\thisfigwidth,clip=true,trim=39 80 23 56]{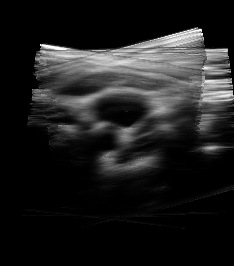}\label{fig:slices-vessel-geomean-lateral}}&
		\subfloat[]{\includegraphics[width=\thisfigwidth,clip=true,trim=39 80 23 56]{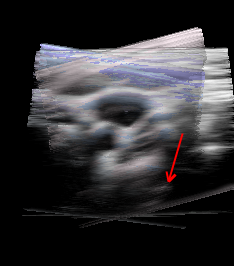}\label{fig:slices-vessel-geomax-lateral}}
		\\[-2em]
		\rotatebox{90}{\centering Carotid ax.}&
		\subfloat[]{\includegraphics[width=\thisfigwidth,clip=true,trim=32 110 96 88]{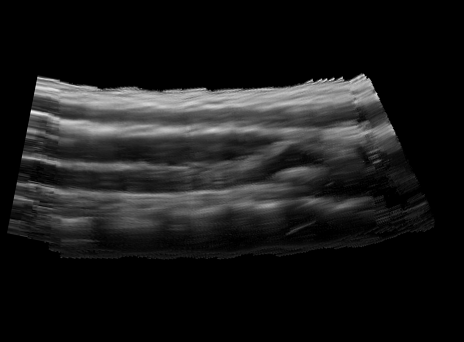}\label{fig:slices-vessel-mean-axial}}&
		\subfloat[]{\includegraphics[width=\thisfigwidth,clip=true,trim=32 110 96 88]{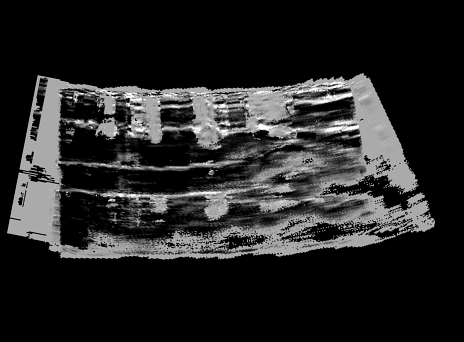}\label{fig:slices-vessel-trace-axial}}&
		\subfloat[]{\includegraphics[width=\thisfigwidth,clip=true,trim=32 110 96 88]{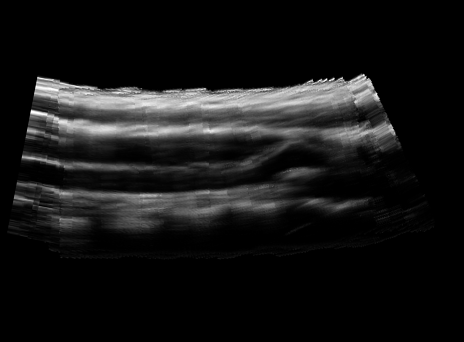}\label{fig:slices-vessel-geomean-axial}}&
		\subfloat[]{\includegraphics[width=\thisfigwidth,clip=true,trim=32 110 96 88]{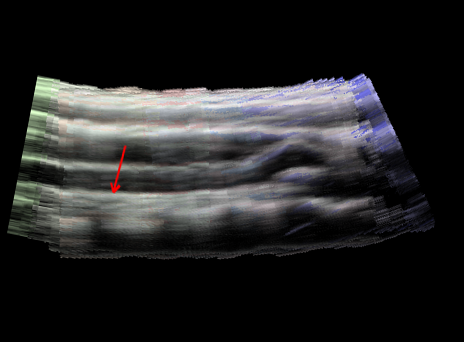}\label{fig:slices-vessel-geomax-axial}}
		\\[-2em]
		\rotatebox{90}{\centering Vessel lat.}&
		\subfloat[]{\includegraphics[width=\thisfigwidth,clip=true,trim=100 75 104 120]{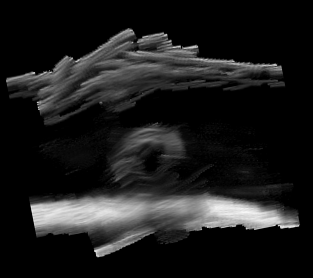}\label{fig:slices-vesselphantom-mean-lateral}}&
		\subfloat[]{\includegraphics[width=\thisfigwidth,clip=true,trim=100 75 104 120]{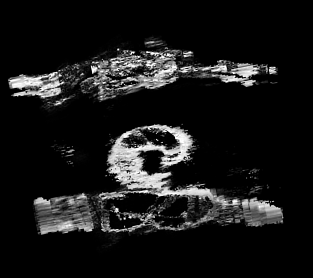}\label{fig:slices-vesselphantom-trace-lateral}}&
		\subfloat[]{\includegraphics[width=\thisfigwidth,clip=true,trim=100 75 104 120]{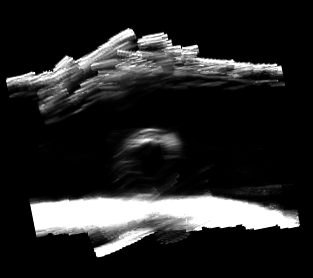}\label{fig:slices-vesselphantom-geomean-lateral}}&
		\subfloat[]{\includegraphics[width=\thisfigwidth,clip=true,trim=100 75 104 120]{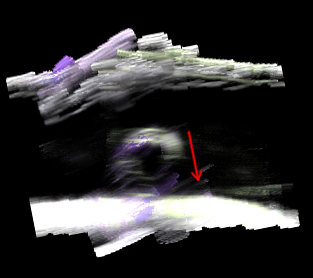}\label{fig:slices-vesselphantom-geomax-lateral}}
		\\[-2em]
		\rotatebox{90}{\centering Vessel ax.}&
		\subfloat[]{\includegraphics[width=\thisfigwidth,clip=true,trim=16 66 22 101]{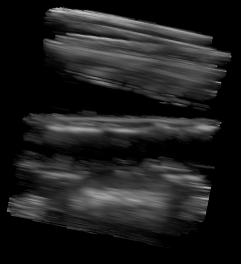}\label{fig:slices-vesselphantom-mean-axial}}&
		\subfloat[]{\includegraphics[width=\thisfigwidth,clip=true,trim=16 66 22 101]{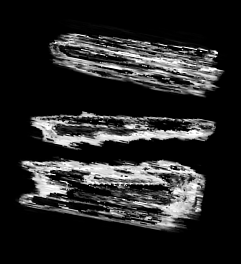}\label{fig:slices-vesselphantom-trace-axial}}&
		\subfloat[]{\includegraphics[width=\thisfigwidth,clip=true,trim=16 66 22 101]{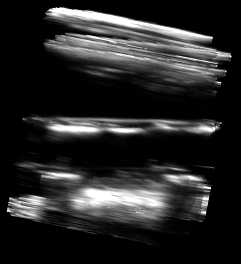}\label{fig:slices-vesselphantom-geomean-axial}}&
		\subfloat[]{\includegraphics[width=\thisfigwidth,clip=true,trim=16 66 22 101]{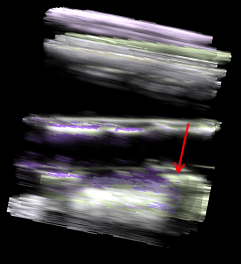}\label{fig:slices-vesselphantom-geomax-axial}}
		\\
		\\[-0.9em]
		& 
		Mean compounding&
		Tensor-CS  &
		Spherical-CS &
		Spherical-CS \\
		&& (Tensor trace) & (Mean of cells)  & (Normal visualization)
	\end{tabular}
	\caption{\shortcaption{3D Reconstructions} from a femoral (Seq. 6, 166 frames, ROI dir.var.~\meanStd{0.0635}{0.0165}) and a carotid (Seq. 5, 437 frames, ROI dir.var.~\meanStd{0.0257}{0.0061}) in-vivo acquisitions as well as from a vessel phantom (Seq. 6, 1166 frames, ROI dir.var.~\meanStd{0.0421}{0.0151}). 
	From left to right we compare the mean reconstruction, the absolute value of the tensor trace in tensor-CS, the mean of all grid-cells in a geodesic grid and in the last column we show the estimated surface normals in color coding (scaled with the geodesic mean).
	The red arrows show the main direction the color coding is based upon.}
	\label{fig:volumes-qualitative}
\end{figure*}

The images reveal first that Tensor-CS has an unstable behavior in unconstrained large scale evaluations.  Given its reduced representation complexity it is prone to fail when the data from different directions is of higher order than our 2nd order model (the intensities vary significantly across different directions in a complex pattern).  With the current implementation this effect is translated into a failure of the tensor fitting, shown as white voxels in the image. This effect seems to be located in boundary areas and locations where numerical instabilities are evoked by low intensities measured from a high number of directions, as can be seen in the phantom views.
Despite this behaviour, it is apparent that whenever the tensor calculation succeeds the details of soft tissues are increased. Moreover, the tissue interfaces, \eg in the two views of the carotid sequence, are best defined in Tensor-CS.
In the case of Spherical-CS, the cell-mean visualization filters artefacts coming from directions with secondary reflections. This can be clearly seen for the femur lateral view: while the mean compounding features reflections behind the femoral interface, these reflections do not appear in the spherical-CS mean reconstruction.
Also, a considerable blur reduction is observed when comparing spherical-CS to the mean compounding, in particular for the two views of the carotid and the lateral vessel sequence.
The color visualization of the maximum Spherical-CS cells provides information about the interfaces' normals, as can be seen in the femoral lateral view but more clearly as the normal of the carotid changes in the carotid lateral view or in the vessel phantom.

\subsection{Free view point evaluation}
\label{subsec:free_view_evaluation}
\begin{figure}
	\centering
	\subfloat[]{
		\includegraphics[clip=true,trim=100 283 108 85,width=0.23\textwidth]{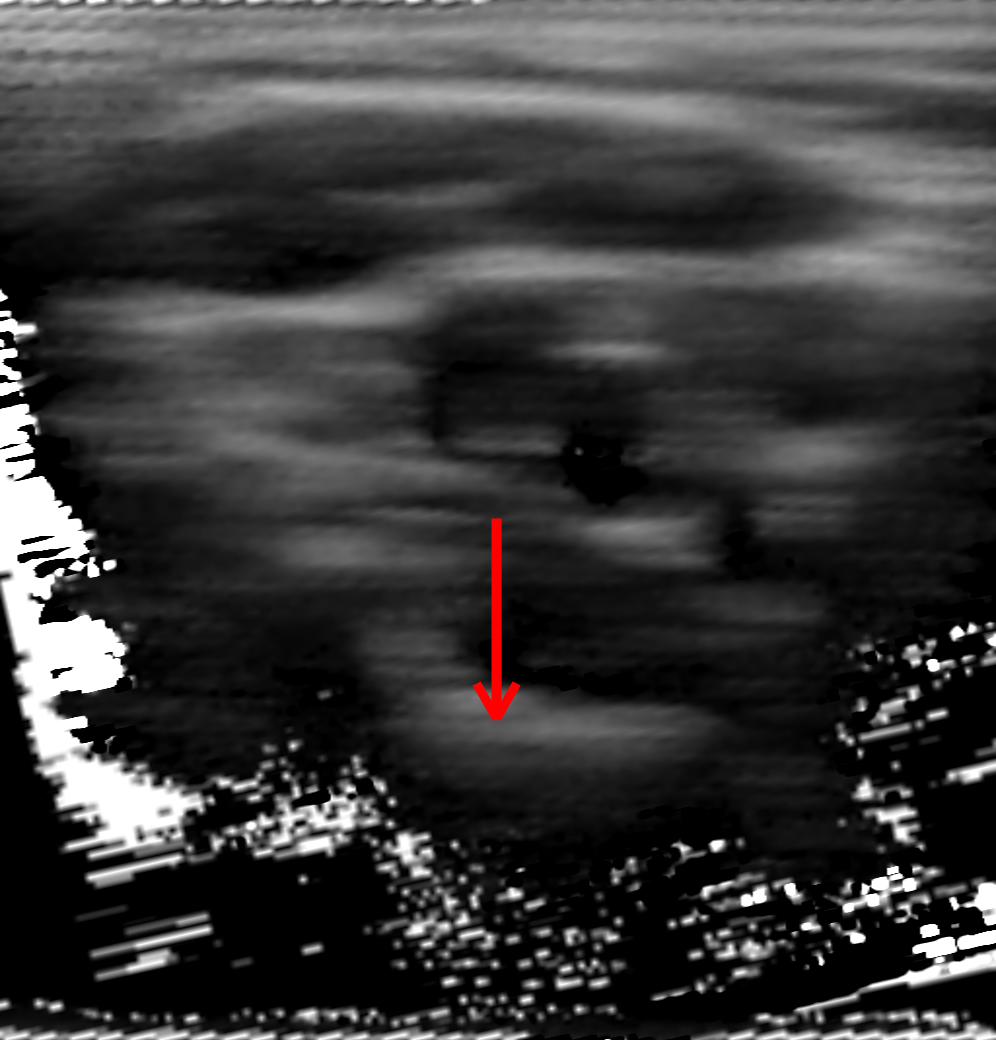}
		\label{fig:free-view-a}}
	\subfloat[]{
		\includegraphics[clip=true,trim=100 283 108 85,width=0.23\textwidth]{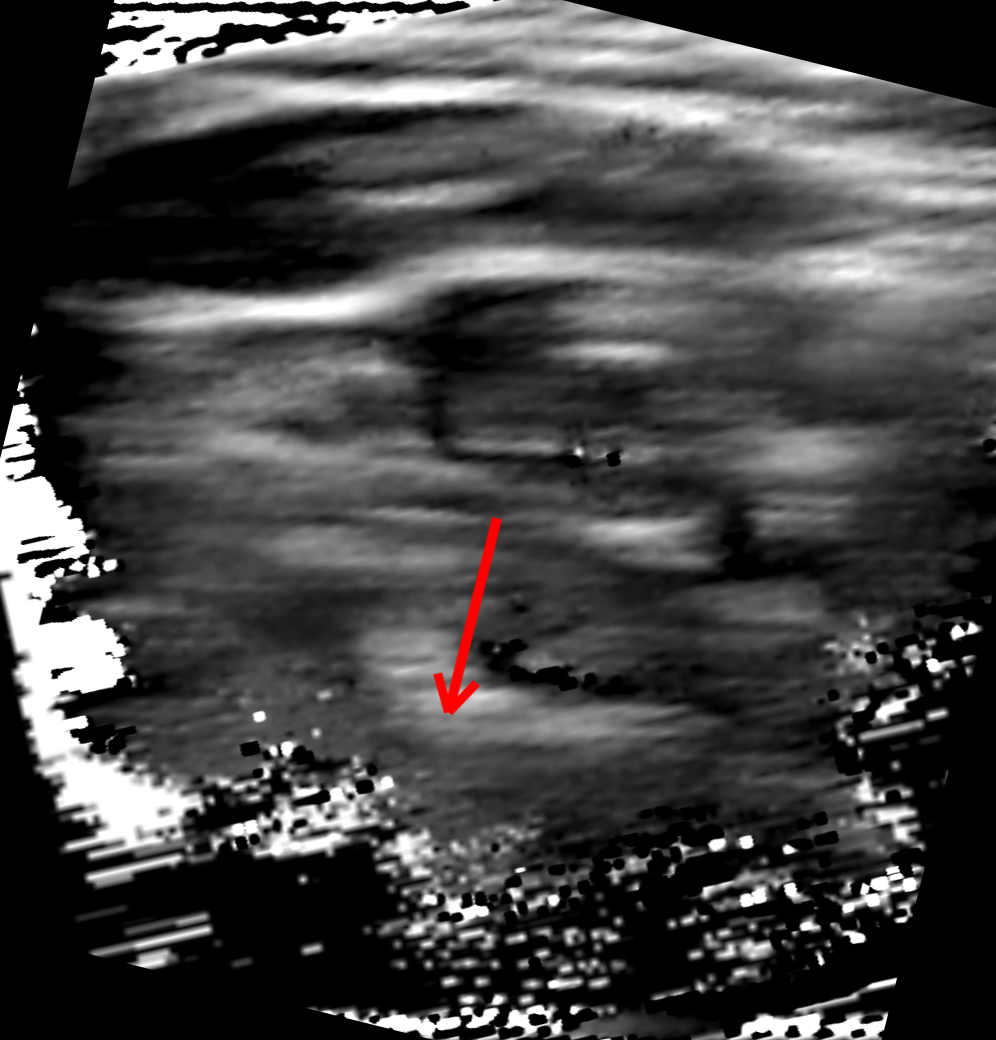}
		\label{fig:free-view-b}}
	\caption{\shortcaption{Tensor-CS based Free-view visualization} on a carotid scan (Seq. 5, 437 frames).
	The red arrows indicate the scanline directions of the reprojections.
	 \mysubref{a} shows a reprojection in original image direction, \mysubref{b} the intensities derived form the tensor model in a different direction.
	 \label{fig:free-view}}
\end{figure}

The directional information preserved in the CS models can be used to estimate intensities from arbitrary directions.
This is achieved by evaluating \eqref{equ:tensorProjection} for tensor-CS with a chosen direction.
\autoref{fig:free-view} shows this for a carotid scan (Seq. 5), where the left is extracted from the original view-point and the image resulting from another direction on the right.
It can be seen that the contrast between the interfaces and the tissue intensities is changed between the two directions.

\section{Discussion}

The qualitative results in combination with the quantitative evaluation show that geodesic spherical CS provides the best representation for arbitrary US-sweeps, while complying with the requirements of data fidelity  and constant storage.
The mappings from samples's directions to the sphere are slightly more complex than for Tensor-CS,  but this is compensated by the fact that as opposed to Tensor-CS, Spherical-CS does not rely on an optimization procedure to reconstruct the cell-values for a given voxel.
As this optimization is based on all samples in a voxel, it can only be performed in a two-step reconstruction as we have presented it.
The spherical-CS on the other hand could be implemented in an iterative reconstruction process, thus allowing for run-times comparable to state of the art scalar reconstructions.
In addition, the current implementation of Tensor-CS has shown an unstable behaviour for general acquisitions with wide spherical variation. We believe that part of the instability could be solved by using more sophisticated optimization algorithms resulting in additional computational cost. However, the inherent limitations of the 2nd order symmetric model would still remain.

The concept of Computational Sonography opens new visualization paths for 3D US, relying on CS ability to store and then compute values on demand. In addition to the visualizations in \autoref{fig:volumes-qualitative} and according to the application, one may want to display the most meaningful cell for each voxel,
to ensure faithfulness to the original measures. 
Online interactive visualization methods can be used to visualize the true directional-dependency of the ultrasound volume even after reconstruction and as such can serve as a basis for realistic simulations for training.
Finally, we may also use the directional information to guide online acquisitions towards a viewpoint where the structure is most visible or to create surfaced visualization of interfaces.

Beyond visualization, Computational Sonography aims at compactly and reliably encoding all the information available during a free-hand acquisition in order to make it available for later processing. For instance, the CS concept is also useful for segmentation where the direction of the maximum cell gives hints about the presence of surfaces. Also for registration, CS allows us enables to compute the better similarity functions between a 2D slice and the reconstructed 3D volume, when accounting for the directional dependency of the images.



\section{Conclusion}
We showed the performance of several methods for the representation of the directional dependent information found in freehand US, using the conventional mean compounding as a baseline. Two storage models were proposed. A tensor-CS model, which we initially proposed in \cite{hennersperger2015computational} and a novel approach based on spherical grids. We illustrated the adaption capabilities of the two representations, showing that the information can be extracted on demand according to the application. 
Finally, an exhaustive quantitative evaluation of the representation error resulted in a favorable performance of the spherical grids over the tensors and mean compounding methods. 
These results serve as evidence to our claim that with our new representation freehand acquisitions do not any longer need to rely on linear protocols.

Future research includes exploring other applications that require more complex algorithms to extract the relevant information, comparable to compressed sensing approaches. Additionally, one could devise methods using non-local information, for example to extract surfaces of tissue interfaces.



\normalsize
\bibliography{components/bib}


\end{document}